\documentclass[journal]{IEEEtran}
\usepackage{amsmath,amsfonts}
\usepackage{algorithm,algpseudocode}
\usepackage{marvosym}

\usepackage{array}
\usepackage[caption=false,font=normalsize,labelfont=sf,textfont=sf]{subfig}
\usepackage{textcomp}
\usepackage{stfloats}
\usepackage{url}
\usepackage{verbatim}
\usepackage{graphicx}
\usepackage[normalem]{ulem}
\usepackage{makecell}   % For the \makecell command
\usepackage{graphicx}   % For the \rotatebox command

\usepackage{cite}
\usepackage{graphicx}
\usepackage{amssymb}
\usepackage{booktabs}
\usepackage{multirow}
\usepackage{xcolor}
\usepackage{colortbl}
\usepackage[capitalize]{cleveref}
\Crefname{section}{Section}{Sections}
\Crefname{table}{Table}{Tables}
\Crefname{figure}{Figure}{Figures}
\definecolor{my_color}{HTML}{e8eef1}
\ifCLASSINFOpdf
\else
\fi
\begin{document}

%\title{Revisiting Task Prior for All-in-One \\Image Restoration with Dynamic Prompts}
\title{Beyond Degradation Redundancy: Contrastive Prompt Learning for All-in-One Image Restoration}

\author{
Gang Wu,~\IEEEmembership{Student Member,~IEEE,}
        ~Junjun Jiang\textsuperscript{\Letter},~\IEEEmembership{Senior Member,~IEEE,}
        ~Kui~Jiang,~\IEEEmembership{Member,~IEEE,}
        ~Xianming~Liu,~\IEEEmembership{Member,~IEEE}
        ~and~Liqiang Nie,~\IEEEmembership{Senior Member,~IEEE}
        
 \thanks{The research was supported by the National Natural Science Foundation of China (62471158, U23B2009) and the Natural Science Foundation of Heilongjiang Province of China for Excellent Youth Project (YQ2024F006).} 
\IEEEcompsocitemizethanks{

\IEEEcompsocthanksitem  G. Wu, J. Jiang, K. Jiang, and X. Liu are with the School of Computer Science and Technology, Harbin Institute of Technology, Harbin 150001, China. E-mail: \{gwu@hit.edu.cn, jiangjunjun@hit.edu.cn, jiangkui@hit.edu.cn, csxm@hit.edu.cn\}. Corresponding author: Junjun Jiang. 
\IEEEcompsocthanksitem L. Nie is with the School of Computer Science and Technology, Harbin Institute of Technology (Shenzhen), Shenzhen 518055, China. E-mail: \{nieliqiang@gmail.com\}.

\IEEEcompsocthanksitem Copyright (c) 2013 IEEE. Personal use of this material is permitted. However, permission to use this material for any other purposes must be obtained from the IEEE by sending a request to pubs-permissions@ieee.org. \protect\\
}
}

\maketitle

\IEEEpeerreviewmaketitle

\begin{abstract}
All-in-One Image Restoration (AiOIR), which addresses diverse degradation types with a unified model, presents significant challenges in designing task-aware prompts that effectively guide restoration across multiple degradation scenarios. While adaptive prompt learning enables end-to-end optimization, it often yields overlapping or redundant task representations. Conversely, explicit prompts derived from pretrained classifiers enhance discriminability but discard critical visual information needed for reconstruction. To address these limitations, we introduce Contrastive Prompt Learning (CPL), a framework that aims to improve prompt-task alignment through two complementary components: a Sparse Prompt Module (SPM) that efficiently captures degradation-aware representations while reducing redundancy, and a Contrastive Prompt Regularization (CPR) that explicitly strengthens task boundaries by incorporating negative prompt samples across different degradation types. Unlike previous approaches that focus primarily on degradation classification, CPL directly optimizes the interaction between prompts and the restoration model. Extensive experiments across five benchmarks show that CPL consistently boosts the performance of strong AiOIR baselines across diverse scenarios. Our approach achieves state-of-the-art average performance on these benchmarks, providing a general and robust solution for AiOIR. The code is available at \url{https://github.com/Aitical/CPLIR}
\end{abstract}


\maketitle
\section{Introduction}
Image restoration (IR) has long been a core problem in low-level computer vision, focusing on recovering high-quality images from degraded observations \cite{Banham1997digitalsurvey, su2022survey}. Common image degradations, such as noise, blur, haze, rain, and poor illumination make image restoration a diverse and specialized field. To address these challenges, researchers have developed tailored solutions for each degradation type, with methods optimized for specific image artifacts \cite{EladKV23DenoiseSurvey, Zhang22DeblurSurvey, chen23DerainSurvey, Gui23DehazeSurvey, Li22LowlightSurvey}. While these methods excel in their respective tasks, their practical utility remains limited. Real-world images are often corrupted by unknown or multiple interacting degradations, demanding a more versatile solution. This has spurred the development of All-in-One Image Restoration (AiOIR) approaches \cite{jiang2024survey}, which strive to handle a broad spectrum of degradations within a single unified model.

\begin{figure}[t]
    \centering
    \includegraphics[width=\linewidth]{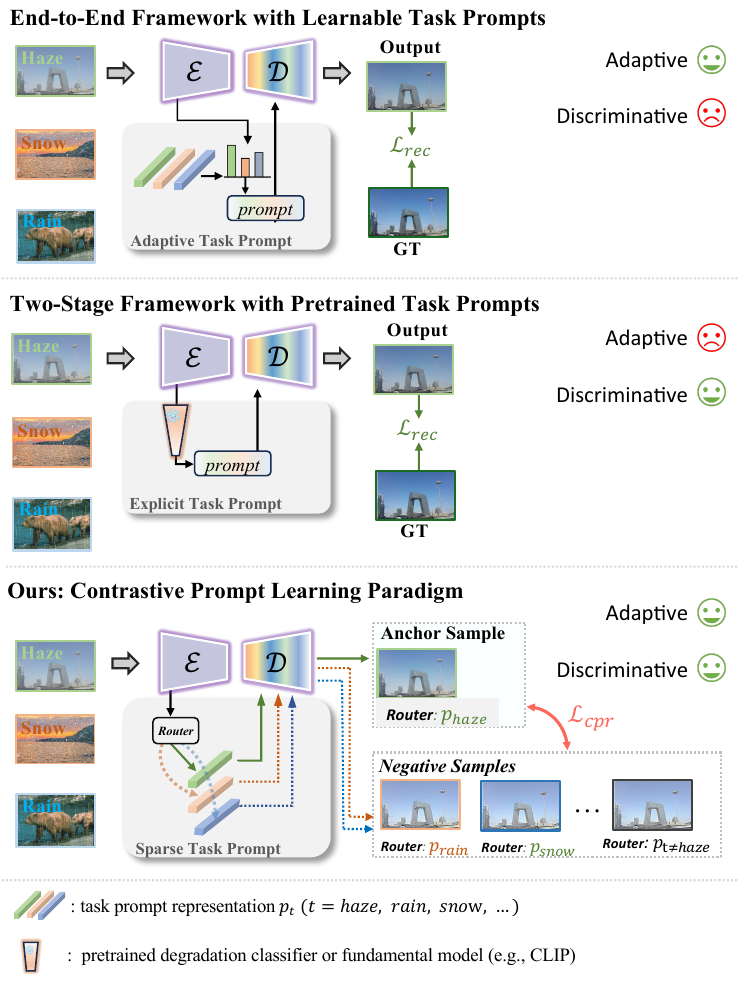}
\caption{Comparison of prompt-based AiOIR frameworks. (a) Adaptive end-to-end learnable prompts. (b) Two-stage frameworks with pretrained degradation encoders. (c) The proposed CPL paradigm with sparse prompt selection and CPR.}

    \label{fig_intro_multi_task}
\end{figure}
 
A common strategy for enabling such versatility is the use of task-aware prompts, which provide the model with additional guidance for tackling different degradations. Research in this area has diverged into two dominant paradigms, each with inherent limitations. The first paradigm, \textit{adaptive prompt learning} \cite{Li22AirNet, Potlapalli2023promptir, cui2025adair}, learns prompts jointly with the restoration network in an end-to-end manner. While offering flexibility, this approach often leads to prompts with high \textit{representation redundancy}, where task-specific information becomes entangled and ambiguous. The second paradigm is \textit{explicit prompt learning} \cite{MioIR, Luo2024DA-CLIP, hu2025promptir_cls}, which leverages pretrained classifiers to generate highly discriminative degradation features. However, this emphasis on discriminability can lead to \textit{functional misalignment}, as features that are optimized for classification are not necessarily optimal for the nuanced task of high-fidelity image reconstruction \cite{Luo2024DA-CLIP}.

\begin{figure*}[t]
    \centering
    \includegraphics[width=0.975\textwidth]{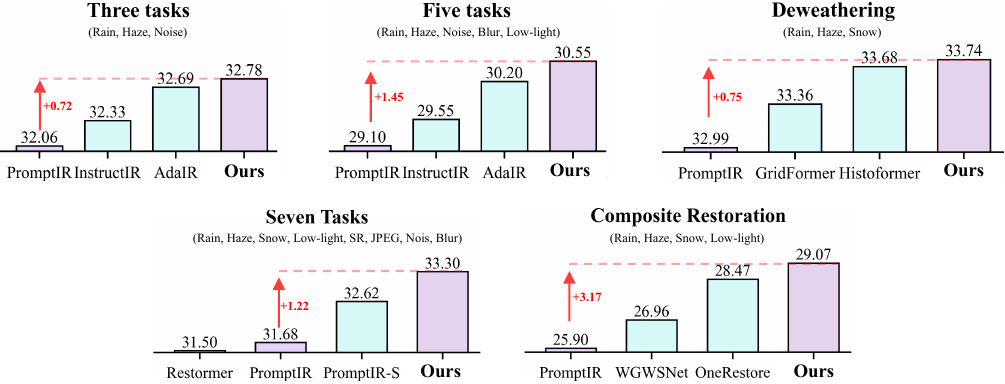}
    \caption{Performance comparison. Integrating our CPL framework into existing all-in-one models improves performance across various tasks in our experiments.}
     
    \label{fig_intro_compare}
\end{figure*}

While these advancements have enabled all-in-one frameworks to handle multiple degradation tasks, several issues remain. Based on existing research \cite{Li22AirNet, Potlapalli2023promptir, MioIR, Luo2024DA-CLIP, hu2025promptir_cls}, task prompts are crucial for AiOIR and are typically obtained through either adaptive or explicit learning. A crucial challenge stems from the inherent commonalities across degradation tasks. For instance, dehazing and low-light enhancement share global illumination patterns, while rain removal and denoising involve similar texture statistics \cite{Kim2022dehazeandlowlight,zheng2019derainanddenoise}. These shared characteristics make it difficult for adaptive learning approaches, where prompts are learned jointly with the restoration model \cite{Li22AirNet,Potlapalli2023promptir, cui2025adair}, to develop clearly separated task representations. This redundancy is quantitatively evident in PromptIR \cite{Potlapalli2023promptir}, which uses a softmax function to ensemble multiple task-oriented prompts and yields a relatively dense probability distribution.
To measure this redundancy, we calculate the Shannon entropy of the softmax probability distribution over different task prompts during inference. Lower entropy values approaching zero indicate that the model clearly distinguishes between tasks, while higher values suggest overlapping representations where multiple prompts are activated simultaneously. Our analysis shows relatively high entropy values: 2.23 bits for rain, 2.08 bits for haze, and approximately 2.24 bits for noise levels 15, 25, and 50, indicating that the learned prompts exhibit limited task specificity. Alternatively, explicit prompt learning approaches employ a two-stage strategy with a pre-trained degradation classifier \cite{MioIR, Luo2024DA-CLIP, hu2025promptir_cls} to yield more discriminative representations. However, recent studies \cite{Luo2024DA-CLIP} have reported that highly discriminative features optimized for classification do not always translate into improved restoration quality, suggesting a misalignment between obtaining discriminative prompts and the requirements of the restoration model. In essence, adaptive methods tend to produce redundant prompts, while explicit methods can produce discriminative but potentially misaligned prompts. This brings us to a central question: \textit{How can we design prompts with sufficient discriminative power while preserving the fine-grained information required by the restoration model?}

\begin{figure*}[t]
    \centering
    \includegraphics[width=\linewidth]{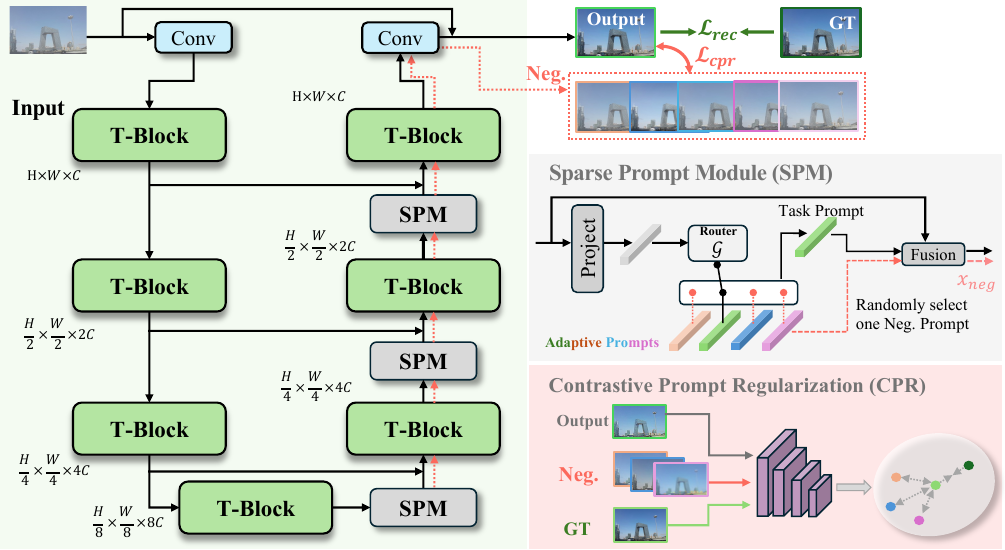}
    \caption{Illustration of the proposed CPL framework with the SPM and CPR. We adopt a stacked Transformer block (T-Block) backbone following \cite{Potlapalli2023promptir}. Red dashed arrows denote the generation of negative samples used only during training.}
    \label{fig_Framework}
\end{figure*}
In this paper, we propose \textit{Contrastive Prompt Learning (CPL)}, a paradigm that addresses the above dilemma through two complementary perspectives. To combat \textit{representation redundancy}, we introduce a SPM. Specifically, we employ a sparse selection mechanism for prompt representations to enhance the intrinsic quality and expressive power of each individual prompt. As illustrated in Figure \ref{fig_intro_multi_task}, it connects adaptive and explicit learning paradigms through a sparse gated router, which derives degradation representations from dense, parameterized prompts. Moreover, to alleviate \textit{functional misalignment}, we propose the CPR. Instead of constraining the prompt embeddings directly, CPR introduces a \textit{model--prompt decoupling strategy} that regularizes the final restoration outcome. By treating mismatched prompt-model pairings as negative samples, CPR directly optimizes the \textit{functional behavior} of the prompts, encouraging them to align with the goal of high-quality reconstruction. This mechanism penalizes the restoration model when prompts are incorrectly assigned to degradations, thereby establishing clearer associations between specific prompt representations and their corresponding degradation types.

By striking a balance between discriminative power and reconstruction quality, our CPL framework reduces cross-task confusion while preserving the information required for high-quality image restoration. As a result, the learned prompts become more task-consistent and better aligned with the underlying degradation characteristics. We conduct extensive experiments on a wide range of AiOIR benchmarks. As illustrated in Figure \ref{fig_intro_compare}, CPL advances the performance of baseline models and attains strong results across all evaluated configurations. Our main contributions are as follows:

\begin{itemize} 

 \item {We reveal and quantify the critical issues of \textit{representation redundancy} and \textit{functional misalignment} in current AiOIR frameworks.}

\item  {We propose a SPM that leverages principled sparsity to enhance the intrinsic quality and expressive power of prompt representations, effectively mitigating redundancy.}

\item  {We introduce CPR, a novel functional regularization paradigm that aligns prompts with the restoration objective through a model-prompt decoupling strategy, a clear departure from standard contrastive learning.}

\item We demonstrate the effectiveness and flexibility of our method through extensive experiments, consistently improving performance across multiple datasets with complex, mixed degradations. 
\end{itemize}

The rest of the paper is organized as follows: Section~\ref{sec:related} reviews related work on both single-task and AiOIR, along with prompt-based learning and contrastive learning methods in computer vision. Section~\ref{sec:method} provides the technical details of our CPL. Section~\ref{sec:experiments} presents experimental results, demonstrating how our framework alleviates prompt misalignment and achieves competitive performance on various benchmarks. Finally, Section~\ref{sec:conclusion} summarizes our main findings and discusses potential future research directions for further enhancing prompt-based AiOIR.

\section{Related Work}
\label{sec:related}
\subsection{Single-Task Image Restoration}
Image restoration is a foundational task in computer vision, aiming to recover high-quality images from degraded inputs affected by various distortions \cite{su2022survey}. With the rapid advancement of deep learning techniques, substantial progress has been achieved in addressing single degradation tasks, where specialized algorithms are developed for specific degradation types, such as image denoising \cite{wan2022old,zhang2023practical}, deblurring \cite{pan2018darkblur,Zhang22DeblurSurvey,Cui2023DualAttention,zamir2022restormer,Quan2024GKMLdeblur}, dehazing \cite{Qin20FFANet, Qin20FFANet,cui2024revitalizing,Feng2024realDehazing,Cui2025EENet}, deraining \cite{jiang2020multi,Jiang2021PCNet,huang2023memroyderain,chen23DerainSurvey,Hsu2023waveletderain,Jiang2025DAWN}, and low-light enhancement \cite{cai2023retinexformer, Li22LowlightSurvey,dai20247k++,Wang2024RFFNet,Zhang2025selfregression}. Beyond these task-specific approaches, several general restoration architectures have emerged that provide strong baseline performance across multiple tasks \cite{chen2022simple,chen2021IPT, Uformer, zamir2022restormer, wang2024gridformer, guo2024mambair,cui2024hybridfrequency,li25mair}, together with diffusion-based architectures \cite{Belhasin2024PUIR,Yue2024Difface,li2024latent,yue2025ResShift}. 
While the above task-specific methods have achieved strong results, their limited ability to generalize across different degradation types restricts their practicality in real-world applications, where images frequently suffer from multiple or unknown degradations simultaneously. To address this limitation, recent research has shifted towards developing AiOIR frameworks capable of handling multiple degradation types within a unified model \cite{jiang2024survey}.

\subsection{All-in-One Image Restoration}
Recent approaches introduce task-aware prompts or conditioning mechanisms to better differentiate between degradation types. AirNet \cite{Li22AirNet} proposed a dedicated degradation encoder to extract degradation-specific features, enabling adaptive restoration across various tasks. Building on this foundation, PromptIR \cite{Potlapalli2023promptir} integrated learnable prompt components during the decoding stage to inject task-related information, improving the model's flexibility and performance. The evolution of AiOIR has proceeded along two primary trajectories. The first focuses on improving learnable task prompts through additional prior regularization techniques, such as frequency analysis in AdaIR \cite{cui2025adair} and principal component analysis in \cite{zhang2023IDR}. These approaches aim to enhance the discriminative power of task representations while maintaining their relevance to restoration objectives.

The second trajectory leverages predefined degradation priors with explicit prompt learning. MioIR \cite{MioIR} introduced a sequential learning strategy, effectively capturing task-specific characteristics through a pretrained degradation classifier. This explicit approach has gained significant attention due to its effectiveness in providing clear degradation signals. Further extending this concept, several studies have incorporated pretrained vision-language models (VLMs) like CLIP \cite{Radford2021CLIP} to inject semantic information into the restoration process \cite{Luo2024DA-CLIP, Lin2024Textual,Conde2024InstructIR}. By utilizing textual descriptions or embeddings, these methods provide additional contextual guidance that aids in reconstructing images with complex degradations. In addition, some works focus on the optimization of AiOIR, introducing multi-task learning approaches \cite{wu2024harmony,wu2025debiased} and masked image pretraining \cite{qin2024maskedIR}. Despite significant advancements, existing all-in-one methods are hindered by two critical issues: representation redundancy and prompt-task misalignment. The lack of explicit constraints in adaptive learning often leads to ambiguous, overlapping prompts that degrade performance. 
To address this, we introduce the SPM, a solution that enhances prompt quality through principled sparse selection. Our approach offers a distinct advantage in its integration efficiency; it can be seamlessly incorporated into existing adaptive frameworks with minimal architectural overhead. Furthermore, while our method shares conceptual similarities with Mixture-of-Experts (MoE) architectures~\cite{Zamfir2025MoCEIR}, the underlying motivation is notably different. Classical MoE primarily employs conditional computation to achieve architectural scalability, i.e., to increase backbone capacity without a proportional increase in computational cost. In contrast, we leverage sparsity as a structural prior for \emph{prompt representation} learning. By enforcing sparse selection among prompt modules, the model is encouraged to let each prompt specialize on a subset of degradations, which in turn reduces overlap between prompts and mitigates the feature redundancy observed in dense prompt ensembles.

\subsection{Contrastive Learning in Low-Level Image Restoration}

Contrastive learning has emerged as a powerful paradigm in self-supervised representation learning, enabling models to learn discriminative features by contrasting positive and negative sample pairs~\cite{Gui2024ContrastiveSurvey}. In recent years, researchers have increasingly adapted contrastive learning principles to address the unique challenges of low-level image restoration tasks~\cite{wu2021contrastive1,Zheng23dehazecontrastive,PCL,Ran2024ilcr_contrastive_derain,Chang2024derainingcontrastive,Gao2024fadformer,wu2024learning}. Unlike high-level vision, where semantic feature discrimination is paramount, restoration tasks require preserving fine-grained details while removing task-specific degradations. This leads to specialized contrastive frameworks across various restoration domains. For image dehazing, Wu \textit{et al.}~       \cite{wu2021contrastive1} pioneered the application of contrastive regularization by treating low-quality inputs as negative samples, while Zheng \textit{et al.}~\cite{Zheng23dehazecontrastive} advanced this approach through progressive negative sample selection within a continual learning framework. In the super-resolution task, Wu \textit{et al.} introduced PCL \cite{PCL}, a practical contrastive learning framework that incorporates hard negative sample construction to better distinguish fine-grained frequency patterns. 

The contrastive paradigm has been further extended to other restoration tasks with domain-specific adaptations. For image deraining, Ran \textit{et al.} \cite{Ran2024ilcr_contrastive_derain} developed contrastive regularization that leverages the decoupled representations of rain masks and background content to enhance separation. Chang \textit{et al.}~\cite{Chang2024derainingcontrastive} proposed an unsupervised deraining method using asymmetric non-local contrastive learning, leveraging intra-layer self-similarity and inter-layer exclusiveness to effectively distinguish rain from clean images without supervision. Gao \textit{et al.} \cite{Gao2024fadformer} proposed FADFormer, which introduces a frequency-aware contrastive loss that operates in the frequency domain to better capture structural degradations. Moving toward more general image restoration frameworks, Wu \textit{et al.} \cite{wu2024learning} introduced a task-agnostic model contrastive learning paradigm that simplifies negative sample construction while maintaining effectiveness across different degradation types. Despite these significant advances, existing contrastive learning approaches in low-level vision remain predominantly designed for single-task scenarios, focusing on enhancing feature representations for specific degradation types. They do not address the unique challenges presented by multi-task or AiOIR frameworks, particularly the potential misalignment issues arising from task-oriented prompts that must seamlessly coordinate multiple restoration objectives simultaneously.

While conventional contrastive methods in low-level vision primarily focus on enhancing feature discrimination within a single degradation context, our work introduces a conceptual shift in applying contrastive learning to this domain. Rather than directly constraining the prompt embeddings, a strategy that risks misaligning them with the reconstruction goal, our CPR targets the functional outcome of the restoration process. The key innovation is a \textit{prompt-model decoupling} strategy, where negative samples are generated from mismatched prompt-model pairings. This directly regularizes the prompt's functional alignment with the model, ensuring it is optimized for reconstruction quality over mere classification.

\begin{figure*}[t]
    \centering
    \includegraphics[width=\linewidth]{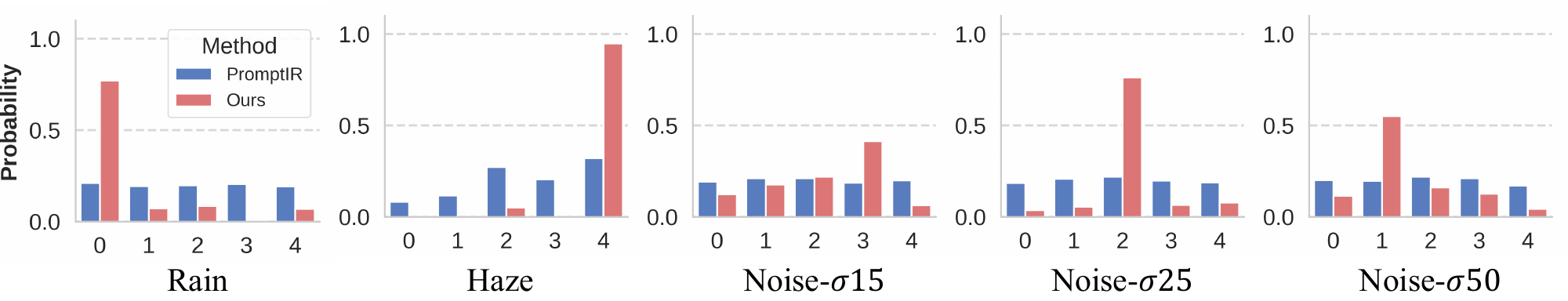}
    \caption{Comparison of prompt selection probability distributions between baseline PromptIR \cite{Potlapalli2023promptir} and CPL across different degradation tasks. The x-axis (0--4) denotes prompt indices and the y-axis denotes selection probabilities. PromptIR (blue) activates multiple prompts for a given task, whereas CPL (red) produces more concentrated, task-specific prompt selections.}

    \label{fig:entropy_analysis}
\end{figure*}

\section{Method}
\label{sec:method}

In this section, we present the Contrastive Prompt Learning (CPL) framework for AiOIR. We address the core challenges of representation redundancy and prompt-task misalignment through two complementary mechanisms. First, the SPM enhances the intrinsic quality of prompts by fostering representation specialization through sparse selection. Second, the \textit{Contrastive Prompt Regularization (CPR)} strategy ensures that these specialized prompts are optimally utilized by enforcing functional alignment between the prompt guidance and the restoration backbone's behavior. The overall framework is illustrated in Figure \ref{fig_Framework}.

\subsection{Preliminaries}
\label{subsec:preliminaries}

AiOIR aims to address various degradation types (e.g., denoising, deblurring, dehazing, deraining, low-light enhancement) within a unified model. Let $t \in \mathcal{T}$ denote a specific degradation type, and let $\mathbf{I}_d$ be the degraded input image. In prompt-based approaches, each task $t$ is associated with a learned prompt $\mathbf{p}_t$ that guides the restoration process. The restoration model, parameterized by $\theta$, can be formulated as:
\begin{equation}
    \mathbf{I}_r 
    = \mathcal{F}\bigl(\mathbf{I}_d,\ \mathbf{p}_t;\ \theta\bigr),
    \label{eq:restoration}
\end{equation}
where $\mathbf{I}_r$ is the restored image. The prompt $\mathbf{p}_t$ should provide distinct, task-specific guidance to remove artifacts associated with task $t$, ensuring that $\mathbf{I}_r$ closely approximates the ground truth $\mathbf{I}_{gt}$.

However, standard adaptive approaches often suffer from \textit{prompt-task misalignment} \cite{Potlapalli2023promptir}, where the learned prompts fail to provide distinct task information. As evidenced by the entropy analysis in Figure~\ref{fig:entropy_analysis}, baseline methods (e.g., PromptIR) exhibit elevated entropy values (2.08--2.24 bits) across degradation types, indicating significant overlap between task representations. This redundancy stems from two limitations: (1) the softmax-based selection mechanism often yields diffuse probability distributions rather than selecting a dominant, task-specific expert; and (2) the optimization objective focuses solely on reconstruction, lacking explicit constraints to enforce prompt discriminability. Our CPL framework addresses these issues directly. As shown in Figure~\ref{fig:entropy_analysis}, our method achieves highly concentrated distributions. By combining sparse prompt selection with contrastive regularization, we establish clearer boundaries between task representations while strengthening their alignment with the restoration process.

\subsection{Sparse Prompt Module}
\label{subsec_sparse_prompt_module}
To mitigate representation redundancy, we introduce the SPM. Unlike conventional approaches that learn dense, entangled prompt representations, SPM employs a dynamic gating mechanism to selectively activate prompts based on the input. This compels each individual prompt expert to develop a more specialized and expressive representation.

\subsubsection{Module Architecture}

The SPM comprises two key components: (1) a set of $n$ learnable prompt experts $\{\mathcal{E}_1, \mathcal{E}_2, ..., \mathcal{E}_n\}$, each encoding distinct degradation priors, and (2) a gating network $\mathcal{G}$ that determines the contribution of each expert. Given input features $\mathbf{x}$ extracted from the degraded image $\mathbf{I}_d$, the SPM generates the prompt $\mathbf{p}$ as follows:
\begin{equation}
    \mathbf{p} = \sum_{i=1}^{n} g_i(\mathbf{x}) \cdot \mathcal{E}_i(\mathbf{x})
    \label{eq:sparse_prompt},
\end{equation}
where $\mathbf{p}$ represents the generated task prompt, $\mathcal{E}_i(\mathbf{x})$ is the output of the $i$-th prompt expert, and $g_i(\mathbf{x})$ is the gating weight for expert $i$ computed by the gating network $\mathcal{G}$.

\subsubsection{Sparse Gating Mechanism}
To enforce specialization, we employ a top-$k$ sparse gating mechanism to activate the most relevant prompts. The gating weights are computed by:
\begin{equation}
    \mathcal{G}(\mathbf{x}) = \text{top-}{k}(\text{softmax}(W_g \mathbf{x} + b_g), k)
    \label{eq:sparse_gating},
\end{equation}
where $W_g$ and $b_g$ are learnable parameters. The $\text{top-}k$ operator retains the $k$ largest values from the softmax distribution and sets the rest to zero, re-normalizing the active weights to sum to one. This sparsity constraint offers two advantages. First, it acts as a regularization term that reduces representational overlap, as prompts must compete to be selected for specific degradation patterns. Second, it improves computational efficiency by activating only a subset of prompts (typically $k \ll n$) during inference, while still allowing the model to learn from a diverse pool of prompts during training.

\subsection{Contrastive Prompt Regularization}
\label{subsec:cpr}

While SPM improves the intrinsic discriminability of the prompts, there remains a risk of \textit{functional misalignment}: a prompt might be distinct in the embedding space but fail to trigger the correct restoration behavior in the IR backbone. To address this, we introduce the CPR. Conventional contrastive learning typically operates on the feature embeddings. However, in low-level restoration, embedding-level discrimination does not guarantee optimal reconstruction. Therefore, CPR shifts the regularization target from the latent representation to the corresponding output of the model. We propose the \textit{model-prompt decoupling concept} that penalizes the model when it produces high-quality results with incorrect prompts, thereby enforcing a strict dependency between the prompt guidance and the restoration outcome.

\begin{algorithm}[t]
\caption{CPR for AiOIR}
\label{alg:CPR}
\begin{algorithmic}[1]
    \Require Training set $\mathcal{D} = \{\bigl(\mathbf{I}_d, \mathbf{I}_{gt}, t\bigr)\}$; prompts $\{\mathbf{p}_t\}$; model $\mathcal{F}(\cdot; \theta)$; hyperparameter $\alpha$.
    \State Initialize model parameters $\theta$.
    \For{each training iteration}
        \State Sample a batch $\{(\mathbf{I}_d, \mathbf{I}_{gt}, t)\}$ from $\mathcal{D}$.
        \For{each sample in the batch}
            \State \textbf{Positive Reconstruction:} 
                $\mathbf{I}_r^{+} 
                = \mathcal{F}\bigl(\mathbf{I}_d,\ \mathbf{p}_t;\ \theta \bigr)$
            \State \textbf{Negative Reconstructions:} 
                \For{each of $m$ randomly chosen prompts $t' \neq t$}
                    \State $\mathbf{I}_r^{-} 
                    = \mathcal{F}\bigl(\mathbf{I}_d,\ \mathbf{p}_{t'};\ \theta \bigr)$
                \EndFor
            \State Compute $\mathcal{L}_{\mathrm{pos}}$ using Eq.~\eqref{eq:pos_loss}
            \State Compute $\mathcal{L}_{\mathrm{neg}}$ using Eq.~\eqref{eq:neg_loss}
            \State $\mathcal{L}_{\mathrm{cpr}} \gets \mathcal{L}_{\mathrm{pos}} 
            - \,\mathcal{L}_{\mathrm{neg}}$
            \State $\mathcal{L}_{\mathrm{pixel}} \gets 
                \|\mathbf{I}_r^{+} - \mathbf{I}_{gt}\|_1$
            \State $\mathcal{L} \gets \mathcal{L}_{\mathrm{pixel}} 
            + \alpha\,\mathcal{L}_{\mathrm{cpr}}$
        \EndFor
        \State Back-propagate and update $\theta$ with $\nabla \mathcal{L}$.
    \EndFor
    \State \Return $\theta$ 
\end{algorithmic}
\end{algorithm}

\subsubsection{Functional Decoupling via Negative Prompting}
For a training sample $(\mathbf{I}_d, \mathbf{I}_{gt})$ associated with task $t$, we construct training pairs defined by their functional validity:
\begin{itemize}
\item \textbf{Positive Pairing (Matched):} The degraded image $\mathbf{I}_d$ is paired with its corresponding prompt $\mathbf{p}_t$. The resulting output $\mathbf{I}_r^{+} = \mathcal{F}(\mathbf{I}_d, \mathbf{p}_t; \theta)$ is optimized to match the ground truth.
\item \textbf{Negative Pairing (Mismatched):} The same image $\mathbf{I}_d$ is paired with a set of randomly sampled mismatched prompts $\mathbf{p}_{t'}$ (where $t' \neq t$). The resulting outputs $\mathbf{I}_{t^{'}}^{-} = \mathcal{F}(\mathbf{I}_d, \mathbf{p}_{t'}; \theta)$ serve as negative samples.
\end{itemize}

The core intuition is that if the restoration model $\mathcal{F}$ is truly prompt-guided, feeding it a ``dehazing'' prompt for a ``rainy'' image should result in a failure to remove the rain (or an incorrect operation), leading to a representation distinct from the ground truth. If the model ignores the prompt and relies solely on image statistics, $\mathbf{I}_{t^{'}}^{-}$ would resemble $\mathbf{I}_r^{+}$, indicating a failure of prompt control.

\begin{figure}[t]
    \centering
    \includegraphics[width=\linewidth]{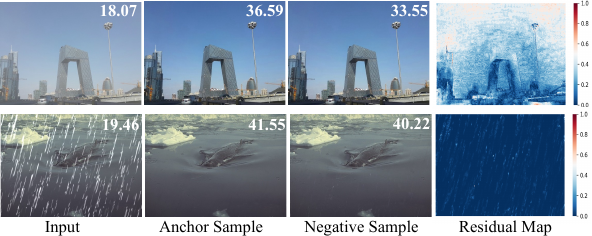}
\caption{Residual analysis for CPR. PSNR values are shown in the top-right corner of each image. Top row: haze example. Bottom row: rain example. Residual maps highlight differences between reconstructions obtained with matched and mismatched prompts.}

    \label{fig:residual_analysis}
\end{figure}

\subsubsection{Contrastive Prompt Loss}

We adopt a pretrained VGG network denoted by $\phi(\cdot)$ to measure perceptual similarity in feature space. Let $\mathbf{I}_{gt}$ be the ground-truth image. We formulate two complementary loss terms:

\paragraph{Positive Loss} This term aims to reduce the distance between the positive reconstruction $\mathbf{I}_r^{+}$ and the ground truth $\mathbf{I}_{gt}$ in feature space:
\begin{equation}
    \mathcal{L}_{\mathrm{pos}}
    = \bigl\|\phi\bigl(\mathbf{I}_r^{+}\bigr) 
           - \phi\bigl(\mathbf{I}_{gt}\bigr)\bigr\|_2^2.
    \label{eq:pos_loss}
\end{equation}

\paragraph{Negative Loss} This term seeks to enlarge the distance between each negative reconstruction $\mathbf{I}_r^{-}$ and $\mathbf{I}_{r}^{+}$:
\begin{equation}
    \mathcal{L}_{\mathrm{neg}}
    = \frac{1}{|\mathcal{T}_{t}^{-}|}
      \sum_{t' \in \mathcal{T}_{t}^{-}}
      \bigl\|\phi\bigl(\mathbf{I}_{t^{'}}^{-}\bigr) 
             - \phi\bigl(\mathbf{I}_{r}^{+}\bigr)\bigr\|_2^2,
    \label{eq:neg_loss}
\end{equation}
where $\mathcal{T}^{-}_t$ denotes the set of $m$ randomly sampled negative task types for the anchor task $t$. We combine these two terms into the \textit{contrastive prompt Regularization loss}:
\begin{equation}
    \mathcal{L}_{\mathrm{cpr}}
    = \mathcal{L}_{\mathrm{pos}}
    - \,\mathcal{L}_{\mathrm{neg}}.
    \label{eq:cp_loss}
\end{equation}

This encourages the model to produce high-quality outputs when using the correct prompt while degrading output quality when using incorrect prompts, effectively strengthening the functional boundary between different task prompts. Algorithm~\ref{alg:CPR} details the training procedure for the proposed CPR approach. For each training sample, we generate both positive and negative reconstructions, compute the relevant losses, and update the model parameters accordingly.

\subsubsection{Overall Training Objective}

In addition to the contrastive prompt loss, we incorporate a standard $\ell_1$ pixel-wise reconstruction term on the \textit{positive} reconstruction:
\begin{equation}
    \mathcal{L}_{\mathrm{pixel}}
    = \bigl\|\mathbf{I}_r^{+} - \mathbf{I}_{gt}\bigr\|_1.
    \label{eq:pixel_loss}
\end{equation}

Our final training objective is:
\begin{equation}
    \mathcal{L}
    = \mathcal{L}_{\mathrm{pixel}}
    + \alpha\,\mathcal{L}_{\mathrm{cpr}},
    \label{eq:total_loss}
\end{equation}
where $\alpha$ is a hyperparameter that controls how strongly the CPR is incorporated. By minimizing this combined loss, the model learns to produce high-quality restorations for matched (positive) prompts while simultaneously establishing clear boundaries between different task representations, effectively addressing both the discriminative power and alignment challenges identified in conventional prompt-based approaches.

\subsection{Remark}\label{subsec:remark}
The effectiveness of CPR is validated by our residual analysis in \Cref{fig:residual_analysis}. While reconstructions from positive and negative pairings may appear visually similar at a glance, the residual maps reveal critical, task-aware differences in their corresponding outcomes. For instance, in the haze example, the residual map highlights significant errors in global illumination patterns when an incorrect prompt is used. Similarly, for rain, the residual map clearly localizes errors to the rain-streak regions. This confirms that CPR successfully conditions the restoration model to be highly sensitive to the prompt's semantic content, ensuring that the prompt acts as a genuine functional switch rather than a passive feature modulator.

\begin{table*}[!htb]
\setlength{\abovecaptionskip}{0pt}
\caption{Quantitative comparison on the three-task AiOIR benchmark. Asterisks (*) indicate results cited from prior work \cite{Potlapalli2023promptir}.}

\label{tab_three_task}
  \centering
\large
\renewcommand\arraystretch{1.15}
    \resizebox{\linewidth}{!}{
    \begin{tabular}{c|l|l|ccc|c|c|c}  
    \toprule[1pt]
 \multirow{2}{*}{\bf Type}& \multirow{2}{*}{\bf Method} &  \multirow{2}{*}{\bf Venue} 
    & \multicolumn{3}{c|}{\textbf{Denoising} (CBSD68\cite{martin2001database1})}
    & \multicolumn{1}{c|}{\bf Dehazing}
    & \multicolumn{1}{c|}{\bf Deraining}
    & \multirow{2}{*}{\bf Average}
 
    \\ \cline{4-8}  % \cmidrule(r){3-5} \cmidrule(r){6-6} \cmidrule(r){7-7}

    & &  & $\sigma = 15$ & $\sigma = 25$ & $\sigma = 50$  
    & SOTS \cite{RESIDE_dataset}
    & Rain100L \cite{yang2017deep}
    &  \\
    \hline
 \multirow{6}{*}{\rotatebox{90}{\textit{General}}} & MPRNet \cite{Zamir_2021_CVPR_mprnet} & CVPR'21
    & 33.27/0.920  & 30.76/0.871  & 27.29/0.761    & 28.00/0.958   & 33.86/0.958  & 30.63/0.940 \\
    & Restormer \cite{zamir2022restormer} & CVPR'22 
    & 33.72/0.930  & 30.67/0.865  & 27.63/0.792    & 27.78/0.958   & 33.78/0.958  & 30.75/0.901 \\
 & NAFNet \cite{chen2022simple} & ECCV'22 
    & 33.03/0.918  & 30.47/0.865  & 27.12/0.754    & 24.11/0.928   & 33.64/0.956  & 29.67/0.844  \\
    & FSNet* \cite{cui2024FSNet}   &  TPAMI'23 
    & 33.81/0.930   &30.84/0.872   & 27.69/0.792     & 29.14/0.968    & 35.61/0.969   & 31.42/0.906   \\
    & DRSformer* \cite{chen2023drsformer}       &    CVPR'23 
    &33.28/0.921   &30.55/0.862   &27.58/0.786     &29.02/0.968    & 35.89/0.970   &31.26/0.902 \\
    & MambaIR* \cite{guo2024mambair} & ECCV'24
    & 33.88/0.931  &  30.95/0.874  & 27.74/0.793    &29.57/0.970   & 35.42/0.969 & 31.51/0.907 \\  
    \hline
 \multirow{9}{*}{\rotatebox{90}{\textit{All-in-One}}} &  DL \cite{dl} & TPAMI'19
    & 33.05/0.914  & 30.41/0.861  & 26.90/0.740    & 26.92/0.391   & 32.62/0.931  & 29.98/0.875 \\
    & AirNet \cite{Li22AirNet} & CVPR'22 
    & 33.92/0.932  & 31.26/0.888  & 28.00/0.797    & 27.94/0.962   & 34.90/0.967  
    & 31.20/0.910  \\
    & IDR* \cite{zhang2023IDR}   & CVPR'23 
    & 33.89/0.931  & 31.32/0.884  & 28.04/0.798    & 29.87/0.970   & 36.03/0.971  & 31.83/0.911  \\

 & Gridformer* \cite{wang2024gridformer}   &  IJCV'24
    & 33.93/0.931  & 31.37/0.887  & 28.11/0.801   
    & 30.37/0.970   & 37.15/0.972  
    & 32.19/0.912  \\
 & NDR \cite{yao2024ndr}   & TIP'24 
    & 34.01/0.932  & 31.36/0.887  & 28.10/0.798    & 28.64/0.962   
    & 35.42/0.969  & 31.51/0.910 \\
 & InstructIR \cite{Conde2024InstructIR}   & ECCV'24 
    & \textbf{34.15}/\textbf{0.933}  & \textbf{31.52}/\textbf{0.890}  & \textbf{28.30}/\textbf{0.804} 
    & 30.22/0.959   & 37.98/0.978  
    & 32.43/0.913 \\
 & TextualDegRemoval\cite{Lin2024Textual} &   CVPR'24
    & 34.01/0.933 & 31.39/0.890  
    & 28.18/0.802  & \textbf{31.63}/\textbf{0.980}    
    & 37.58/0.979  & 32.63/0.917 \\
& AdaIR \cite{cui2025adair}
& ICLR'25
& \underline{34.12/0.935}  
& 31.45/0.892  
& 28.19/0.802  
& 31.06/0.980  
& \underline{38.64/0.983}
& \underline{32.69/0.918}
 \\
\cline{2-9}
  & PromptIR \cite{Potlapalli2023promptir}    & NeurIPS'23
    & 33.98/0.933  & 31.31/0.888  & 28.06/0.799   
    & 30.58/0.974   & 36.37/0.972  
    & 32.06/0.913  \\
    & \cellcolor{my_color}\textbf{PromptIR+CPL} 
    & \cellcolor{my_color}2025 
& \cellcolor{my_color}\textbf{34.15}/\textbf{0.933}
& \cellcolor{my_color}\underline{31.50/0.889}
& \cellcolor{my_color}\underline{28.23/0.800}
& \cellcolor{my_color}\underline{31.27/0.980}
& \cellcolor{my_color}\textbf{38.77}/\textbf{0.985}
& \cellcolor{my_color}\textbf{32.78}/\textbf{0.917} \\
  \bottomrule[1pt]
  \end{tabular} 
  }\vspace{-2mm}
\end{table*}
\begin{figure*}[!h]
    \centering
    \includegraphics[width=\linewidth]{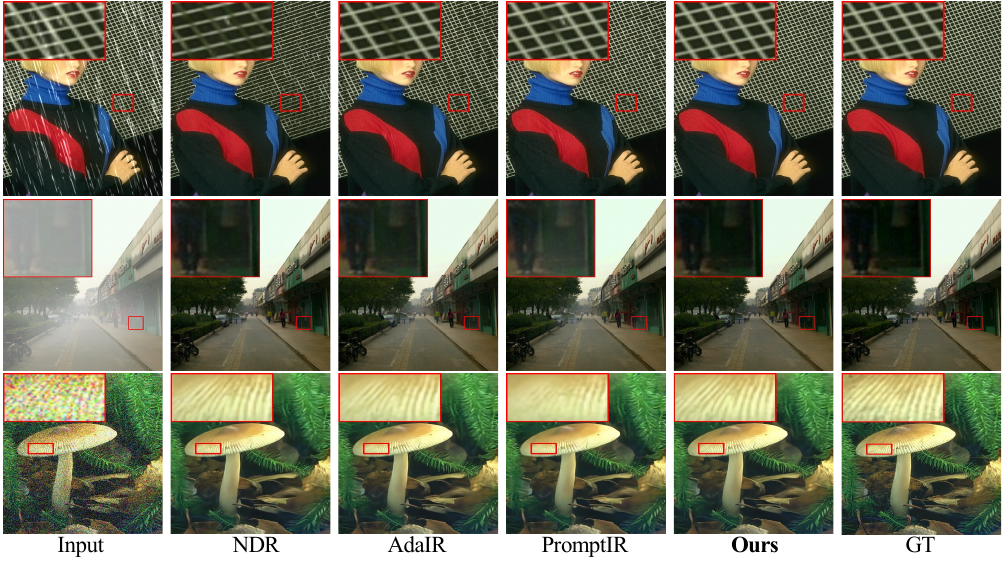}
\caption{Visual comparison on the three-task benchmark. CPL produces clearer structures, fewer residual artifacts, and better detail preservation compared with prior AiOIR methods.}

    \label{fig_3task}
\end{figure*}

\begin{table*}[tb]
\setlength{\abovecaptionskip}{0pt}
\caption{Quantitative comparison on the five-task AiOIR benchmark. Asterisks (*) indicate results cited from prior work \cite{Potlapalli2023promptir,yao2024ndr}.}

  \label{tab_five_task}
  \centering
  \large
  \renewcommand\arraystretch{1.2}
    \resizebox{\linewidth}{!}{
  \begin{tabular}{c|l|l|c|c|c|c|c|c} 
   
    \toprule[1pt]
 \multirow{2}{*}{\bf Type} & \multirow{2}{*}{\bf Method} & \multirow{2}{*}{\bf Venue} 
    & \multicolumn{1}{c|}{\bf Denoising}
    & \multicolumn{1}{c|}{\bf Dehazing}
    & \multicolumn{1}{c|}{\bf Deraining}
    & \multicolumn{1}{c|}{\bf Deblurring}
    & \multicolumn{1}{c|}{\bf Low-light}
    & \multirow{2}{*}{\bf Average}
    \\ \cline{4-8} 
 &   & & CBSD68 \cite{martin2001database1} & SOTS \cite{RESIDE_dataset} & Rain100L \cite{yang2017deep} & GoPro \cite{nah2017deep} & LOL \cite{wei2018deep} &
   \\
    \hline
   \multirow{8}{*}{\rotatebox{90}{\textit{General}}}   &  SwinIR \cite{SwinIR}&  ICCVW'21 
    & 30.59/0.868  & 21.50/0.891  & 30.78/0.923    & 24.52/0.773   & 17.81/0.723  & 25.04/0.835 
  \\
 &Restormer \cite{zamir2022restormer} & CVPR'22 
    & \underline{31.49}/0.884  & 24.09/0.927  & 34.81/0.962    & 27.22/0.829   & 20.41/0.806  & 27.60/0.881  \\

 &NAFNet \cite{chen2022simple} &  ECCV'22 
    & 31.02/0.883  & 25.23/0.939  & 35.56/0.967    & 26.53/0.808   & 20.49/0.809  & 27.76/0.881  \\
 & DRSformer* \cite{chen2023drsformer}&  CVPR'23 
    &30.97/0.881   &24.66/0.931   &33.45/0.953     &25.56/0.780    &21.77/0.821   & 27.28/0.873
 \\
 & Retinexformer* \cite{cai2023retinexformer}   & ICCV'23  
    & 30.84/0.880  &24.81/0.933   &32.68/0.940     &25.09/0.779    &  22.76/0.834   & 27.24/0.873
    \\
 &FSNet* \cite{cui2024FSNet}&  TPAMI'23  
    & 31.33/0.883  &  25.53/0.943  & 36.07/0.968    & 28.32/0.869   & 22.29/0.829  &  28.71/0.898 \\
 &MambaIR* \cite{guo2024mambair} & ECCV'24 
    & 31.41/0.884  &  25.81/0.944  & 36.55/0.971    & 28.61/0.875   & 22.49/0.832  & 28.97/0.901  \\

    \hline
     
   \multirow{9}{*}{\rotatebox{90}{\textit{All-in-One}}} &    DL \cite{dl} & TPAMI'19 
    & 23.09/0.745  & 20.54/0.826  & 21.96/0.762    & 19.86/0.672   & 19.83/0.712  & 21.05/0.743
  \\

 &  TAPE \cite{TAPE} & ECCV'22 
    & 30.18/0.855  & 22.16/0.861  & 29.67/0.904    & 24.47/0.763   & 18.97/0.621  & 25.09/0.801
 \\

   &  Transweather \cite{Transweather}&  CVPR'22 
    & 29.00/0.841  & 21.32/0.885  & 29.43/0.905    & 25.12/0.757   & 21.21/0.792  & 25.22/0.836
 \\

   & AirNet \cite{Li22AirNet} &   CVPR'22 
    & 30.91/0.882  & 21.04/0.884  & 32.98/0.951    & 24.35/0.781   & 18.18/0.735  & 25.49/0.846
 \\

   &  IDR \cite{zhang2023IDR}&  CVPR'23 
    & \textbf{31.60}/0.887  & 25.24/0.943  & 35.63/0.965    & 27.87/0.846   & 21.34/0.826  & 28.34/0.893  \\

 &  Gridformer* \cite{wang2024gridformer} &  IJCV'24 
    & 31.45/0.885  & 26.79/0.951  & 36.61/0.971   
    & \underline{29.22}/0.884   & 22.59/0.831  
    & 29.33/0.904 
 \\ 

 & InstructIR \cite{Conde2024InstructIR} &  ECCV'24 
    & 31.40/0.887  & 27.10/0.956  & 36.84/0.973    
    &  \textbf{29.40}/0.886   &  \underline{23.00}/0.836  
    & 29.55/0.907 
  \\
 & AdaIR\cite{cui2025adair}
 & ICLR'25
 & 31.35/0.889 
 & \underline{30.53}/0.978
 & \underline{38.02}/0.981 
 & 28.12/0.858 
 & \underline{23.00}/0.845 
 & \underline{30.20}/0.910
  \\
\cline{2-9}
  & PromptIR* \cite{Potlapalli2023promptir} 
  &  NeurIPS'23 
    & 31.47/0.886  
    & 26.54/0.949  
    & 36.37/0.970   
    & 28.71/0.881   
    & 22.68/0.832  
    & 29.15/0.904 \\

 &\cellcolor{my_color}\textbf{PromptIR+CPL} 
 &\cellcolor{my_color}2025
 &\cellcolor{my_color}31.41/0.887  
 &\cellcolor{my_color}\textbf{30.82}/\textbf{0.978}
 &\cellcolor{my_color}\textbf{38.20}/\textbf{0.983}   
 &\cellcolor{my_color}28.72/0.874   
 &\cellcolor{my_color}\textbf{23.65}/\textbf{0.855} 
 &\cellcolor{my_color}\textbf{30.55}/\textbf{0.915} \\
     
  \bottomrule[1pt]
  \end{tabular}   
}\vspace{-4mm}
\end{table*}

\section{Experiments} \label{sec:experiments}
We evaluate the CPL framework across diverse AiOIR settings. We first outline the experimental setup and implementation details (Sec.~\ref{subsec:exp_setup}, \ref{subsec:implementation}), followed by comparisons with state-of-the-art methods (Sec.~\ref{subsec:comparison_sota}). We then conduct ablation studies (Sec.~\ref{subsec:ablation}) and conclude with limitations and future work (Sec.~\ref{subsec:discussion}).

\subsection{Experimental Setup} \label{subsec:exp_setup}

We evaluate CPL across several settings that cover both synthetic and real-world AiOIR scenarios:

\begin{enumerate} 

    \item \textbf{Three-task setting.} Following \cite{Li22AirNet,Potlapalli2023promptir}, we evaluate image denoising, deraining, and dehazing. For denoising, we train on BSD400 \cite{BSD500} and WED \cite{WED_dataset} and test on CBSD68 \cite{martin2001database1} at noise levels $\sigma = 15, 25, 50$. For deraining, we use Rain100L \cite{yang2017deep}, and for dehazing, we adopt the SOTS outdoor set from RESIDE \cite{RESIDE_dataset}.

    \item \textbf{Five-task setting.} To evaluate broader task diversity, we follow \cite{zhang2023IDR} and consider denoising (CBSD68), dehazing (SOTS), deraining (Rain100L), motion deblurring (GoPro \cite{nah2017deep}), and low-light enhancement (LOL \cite{wei2018deep}). This configuration additionally includes motion deblurring and low-light enhancement compared to the three-task setting.

    \item \textbf{Seven-task setting.} We adopt the protocol of \cite{MioIR}, where images are synthesized from DF2K using seven degradations: super-resolution, blur, noise, JPEG compression, rain, haze, and low-light. This setting stresses the ability to handle many heterogeneous degradations within a single model.

    \item \textbf{All-weather deweathering.} Following \cite{Transweather}, we train on the AllWeather dataset \cite{Transweather}, which contains samples from Raindrop \cite{raindrop}, Outdoor-Rain \cite{outdoor}, and Snow100K \cite{Snow100K_dataset}. This setting evaluates AiOIR under diverse synthetic weather conditions.

    \item \textbf{Real-world deweathering.} To assess generalization beyond synthetic data, we further evaluate on WeatherBench \cite{weatherbench_data}, a large-scale benchmark with 41,402 training and 600 test images captured under diverse real-world weather conditions (rain, haze, snow, and combinations thereof).

    \item \textbf{Composite degradation setting.} To evaluate the model's generalization across complex degradation scenarios, we follow \cite{guo2024onerestore} and utilize the CDD-11 dataset, which is explicitly designed for composite degradations. CDD-11 is synthesized by applying eleven types of degradation (including haze, rain, snow, low-light, and their combinations) onto high-quality clear images using physically based rendering. 

\end{enumerate}

For all settings, we adopt baseline methods using the same training/testing splits and protocols as in the original works to ensure a fair comparison. We then integrate the proposed CPL framework into these baselines to highlight the performance improvements driven by the CPL strategy. Unless otherwise specified, the best result in result table is highlighted in \textbf{bold}, and the second best is \underline{underlined}.

\subsection{Implementation Details}
\label{subsec:implementation}

\paragraph{Training Details} All models are implemented in PyTorch and trained on two NVIDIA RTX 3090 GPUs. Unless otherwise specified, we use the Adam optimizer with an initial learning rate of $2 \times 10^{-4}$. We use a batch size of 8 and randomly crop $128 \times 128$ patches for training. Standard data augmentation, including random horizontal/vertical flips and $90^\circ$ rotations, is applied.

\paragraph{CPL Integration} Specifically, we take PromptIR \cite{Potlapalli2023promptir} as the default baseline and replace its prompt-selection component with our SPM. For each training sample, we generate multiple negative reconstructions using mismatched prompts to compute the CPR term. The VGG network used in CPR is kept fixed. The loss weight $\alpha$ in Eq.~\eqref{eq:total_loss} is set to $0.01$, unless otherwise noted in the ablation studies. For SPM, we use a top-$k$ sparse gating mechanism. The number of active experts $k$ is set to 1 by default, based on the ablation results in Sec.~\ref{subsec:ablation}. During inference, only the selected experts contribute to the prompt, so the computational overhead is marginal.
%%%%%%%%%%%%

\begin{table*}[t]
    \centering
\caption{Quantitative comparison on the seven-task AiOIR benchmark. PromptIR variants are based on the Restormer backbone with different training strategies: sequential learning (S), explicit prompt learning (EP), and the proposed CPL.}

\label{table_7_task}
\small
\renewcommand\arraystretch{1.2}
    
\resizebox{0.8\linewidth}{!}{
\begin{tabular}{l|cccccccc}
\toprule[1pt]
\textbf{Method} & \textbf{SR} & \textbf{Blur} & \textbf{Noise} & \textbf{JPEG} & \textbf{Rain} & \textbf{Haze} & \textbf{Low-Light} & \textbf{Avg.} \\
\hline 
SwinIR-S & 26.02 & 31.58 & \textbf{31.36} & 33.40 & 36.69 & \textbf{29.58} & 34.64 & 31.90 \\
Uformer-S & 26.07 & 31.11 & 30.96 & 33.27 & 35.96 & \underline{28.29} & 32.80 & 31.21 \\
Restormer-S & 25.95 & 31.55 & 30.86 & 33.24 & 38.06 & 25.48 & 36.69 & 31.69 \\
Restormer+EP & 25.82 & 31.87 & 30.94 & 33.17 & 36.22 & 26.60 & \underline{40.37} & 32.14 \\
PromptIR & 25.86 & 31.46 & 30.75 & 33.07 & 35.76 & 26.62 & 39.62 & 31.88 \\
PromptIR-S & \underline{26.14} & \underline{32.02} & 31.08 & \underline{33.43} & \underline{39.97} & 27.21 & 38.46 & \underline{32.62} \\
\cellcolor{my_color}\textbf{PromptIR+CPL} 
& \cellcolor{my_color}\textbf{26.22} 
& \cellcolor{my_color}\textbf{32.24} 
& \cellcolor{my_color}\underline{31.28} 
& \cellcolor{my_color}\textbf{33.61} 
& \cellcolor{my_color}\textbf{40.91} 
& \cellcolor{my_color}\underline{28.29}
& \cellcolor{my_color}\textbf{40.55}
& \cellcolor{my_color}\textbf{33.30}  \\
\bottomrule[1pt]
\end{tabular}
}
\end{table*}

\subsection{Comparison with State-of-the-Art Methods}
\label{subsec:comparison_sota}

\paragraph{Three-Task All-in-One Degradation Experiment}
Following established protocols \cite{Li22AirNet,Potlapalli2023promptir}, we evaluate performance on denoising, deraining, and dehazing. Table \ref{tab_three_task} presents comparisons against specialized and all-in-one approaches. The results demonstrate that CPL delivers consistent performance improvements when applied to existing architectures. Several key observations emerge: First, our model obtains an average PSNR of 32.78 dB, which is slightly higher than AdaIR \cite{cui2025adair} (32.69 dB). This gain suggests that our decoupling strategy effectively mitigates task interference that is common in all-in-one frameworks. Second, gains are pronounced for spatially complex degradations. For deraining, our approach achieves 38.77 dB PSNR (0.13 dB higher than AdaIR and 2.40 dB above the baseline PromptIR), while for dehazing, it delivers 31.27 dB. These improvements underscore the capability of CPL to capture spatially varying characteristics through adaptive feature extraction. Furthermore, Figure \ref{fig_3task} provides visual comparisons. For denoising (bottom row), PromptIR+CPL recovers texture details (e.g., zebra stripes) while suppressing artifacts. In dehazing (middle row) and deraining (top row), our approach achieves superior color fidelity and structural preservation compared to baseline methods.

\paragraph{Five-Task All-in-One Degradation Experiment}
To assess scalability, we extend the evaluation to a more diverse five-task benchmark, incorporating motion deblurring and low-light enhancement. As shown in Table \ref{tab_five_task}, our model achieves an average PSNR of 30.55 dB, outperforming the recent state-of-the-art AdaIR by 0.35 dB. A critical observation is the method's robustness in handling distinct degradation mechanisms. Dehazing (veiling effect) and low-light enhancement (illumination recovery) involve global property shifts that differ significantly from additive noise removal. Relative to the PromptIR baseline, we observe improvements of +4.28 dB in dehazing and +0.97 dB in low-light enhancement, which can be attributed to our synergistic design. Specifically, the sparse prompt mechanism encourages individual prompts to acquire richer and more expressive representations by enforcing selection competition. In conjunction with CPR, this approach effectively decouples task-specific features while enhancing the degradation-aware information essential for the restoration model. This ensures that the model receives precise, non-conflicting guidance, thereby enabling it to accommodate diverse restoration objectives within a unified framework.

\begin{table*}[!t]
\setlength{\abovecaptionskip}{5pt}
\setlength{\belowcaptionskip}{0pt}
\small
\renewcommand\arraystretch{1} 
    \centering
    \caption{Quantitative comparisons for adverse weather removal. Methods capable of handling multiple degradation tasks are listed together, and their average performance is provided at the bottom. Missing values are denoted by '--'.}
    \label{tab_allweather} 
\large
    
\setlength{\tabcolsep}{4pt}
 \renewcommand\arraystretch{1.2}
    \resizebox{\linewidth}{!}{
\begin{tabular}{c|l|l|cc|cc|cc|cc|cc}
    \toprule[1pt]
    \multirow{2}{*}{\textbf{Type}} & \multirow{2}{*}{\textbf{Method}} &  \multirow{2}{*}{\textbf{Venue}}   &  
    \multicolumn{2}{c|}{\textbf{Snow100K-S}~\cite{Snow100K_dataset}} & \multicolumn{2}{c|}{\textbf{Snow100K-L}~\cite{Snow100K_dataset}} & 
    \multicolumn{2}{c|}{\textbf{Outdoor-Rain}~\cite{outdoor}} & \multicolumn{2}{c|}{\textbf{RainDrop}~\cite{raindrop}} & \multicolumn{2}{c}{\textbf{Average}} \\
    
    \cline{4-13}
    & & & PSNR & SSIM & PSNR & SSIM & PSNR & SSIM & PSNR & SSIM & PSNR & SSIM \\
        \hline
   \multirow{8}{*}{\rotatebox{90}{\textit{Task-Specific}}} 
   &  SPANet~\cite{SPANet_WangY0C0L19}   & CVPR'19    & 29.92 & 0.8260 & 23.70 & 0.7930 & -- & -- & -- & -- & -- & -- \\
     &    DesnowNet~\cite{Snow100K_dataset}    & TIP'18  & 32.33 & 0.9500 & 27.17 & 0.8983 & -- & -- & -- & -- & -- & -- \\

     & HRGAN~\cite{outdoor}      &   CVPR'19  & -- & -- & -- & -- & 21.56 & 0.8550 & -- & -- & -- & -- \\
     &   MPRNet~\cite{Zamir_2021_CVPR_mprnet} &  CVPR'21 &     -- & -- & -- & -- & 28.03 & 0.9192 & -- & -- & -- & -- \\
     &  AttentiveGAN~\cite{raindrop}       & CVPR'18 & -- & -- & -- & -- & -- & -- & 31.59 & 0.9170 & -- & -- \\
     &    IDT~\cite{xiao2022image} & TIP'22  &     -- & -- & -- & -- & -- & -- & 31.87 & 0.9313 & -- & -- \\
     &   NAFNet~\cite{chen2022simple} &   ECCV'22  &  34.79 & 0.9497 & 30.06 & 0.9017 & 29.59 & 0.9027 & -- & -- & -- & -- \\
     &  Restormer~\cite{zamir2022restormer}       & CVPR'22 & 36.01 & 0.9579 & 30.36 & 0.9068 & 30.03 & 0.9215 & 32.18 & 0.9408 & -- & -- \\
        \hline
    \multirow{12}{*}{\rotatebox{90}{\textit{All-in-One}}} 
     & All-in-One~\cite{as2020}   & CVPR'20  & -- & -- & 28.33 & 0.8820 & 24.71 & 0.8980 & 31.12 & 0.9268 & 28.05 & 0.9023 \\
     &   Transweather~\cite{Transweather}    & CVPR'22   & 32.51 & 0.9341 & 29.31 & 0.8879 & 28.83 & 0.9000 & 30.17 & 0.9157 & 30.20 & 0.9094 \\
     &  WGWSNet~\cite{weather_data}    &  CVPR'22  & 34.31 & 0.9460 & 30.16 & 0.9007 & 29.32 & 0.9207 & 32.38 & 0.9378 & 31.54 & 0.9263 \\
       &    WeatherDiff\(_{64}\)~\cite{ozan2023weatherdiff}   & TPAMI'23 & 35.83 & 0.9566 & 30.09 & 0.9041 & 29.64 & 0.9312 & 30.71 & 0.9312 & 31.57 & 0.9308 \\
       & WeatherDiff\(_{128}\)~\cite{ozan2023weatherdiff}       & TPAMI'23 & 35.02 & 0.9516 & 29.58 & 0.8941 & 29.72 & 0.9216 & 29.66 & 0.9225 & 31.00 & 0.9225 \\
       &    AWRCP~\cite{AWRCP_YeCBSXJYCL23}  & ICCV'23  & 36.92 & 0.9652 & 31.92 & \textbf{0.9341} & 31.39 & 0.9329 & 31.93 & 0.9314 & 33.04 & \underline{0.9409} \\
    & GridFormer~\cite{wang2024gridformer}     & IJCV'24 & \underline{37.46} & 0.9640 & 31.71 & 0.9231 & 31.87 & 0.9335 & 32.39 & 0.9362 & 33.36 & 0.9392 \\
    &  MPerceiver~\cite{AiHZW024}  & CVPR'24  & 36.23 & 0.9571 & 31.02 & 0.9164 & 31.25 & 0.9246 & \textbf{33.21} & 0.9294 & 32.93 & 0.9319 \\
    &  DTPM~\cite{DTPM_0001CCXQL024}  & CVPR'24  & 37.01&  \underline{0.9663} & 30.92 & 0.9174 &  30.99   &  0.9340 & 32.72 & \underline{0.9440} & 32.91 & 0.9404\\
    &  Histoformer~\cite{Histoformer_SunRGWC24} & ECCV'24 & 37.41 & 0.9656 & \underline{32.16} & 0.9261 & \underline{32.08} & \underline{0.9389} & \underline{33.06} & \textbf{0.9441} & \underline{33.68} & \textbf{0.9437} \\
    \cline{2-13}
     &  PromptIR~\cite{Potlapalli2023promptir} & NeurIPS'23 & 36.88 & 0.9643 
     & 31.34 & 0.9200 
     & 30.80 & 0.9229 
     & 32.20 & 0.9359 
     & 32.80 & 0.9357 \\
     
    &  \cellcolor{my_color}\textbf{PromptIR+CPL}   & \cellcolor{my_color}2025 
    & \cellcolor{my_color}\textbf{37.82} & \cellcolor{my_color}\textbf{0.9667} 
    & \cellcolor{my_color}\textbf{32.27} & \cellcolor{my_color}\underline{0.9280} 
    & \cellcolor{my_color}\textbf{32.16} & \cellcolor{my_color}\textbf{0.9417} 
    & \cellcolor{my_color}32.73 & \cellcolor{my_color}0.9428 
    & \cellcolor{my_color}\textbf{33.74} & \cellcolor{my_color}\textbf{0.9437} \\
        \bottomrule[1pt]
    \end{tabular}%
    } 
\end{table*}
\begin{figure*}[!h]
    \centering
    \includegraphics[width=\textwidth]{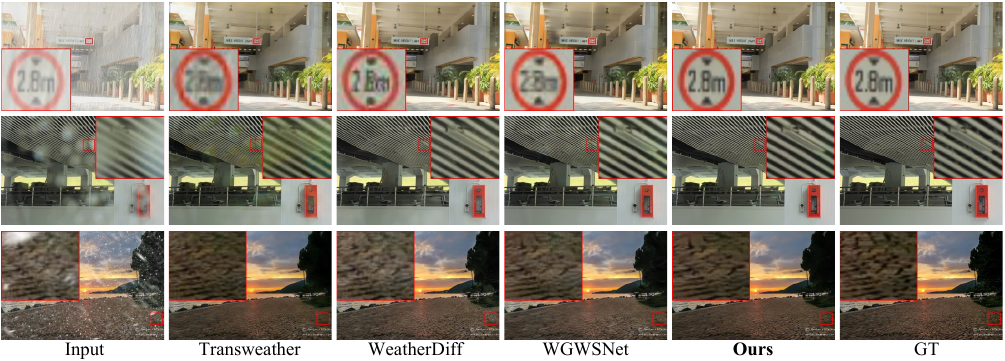}
\caption{Visual comparisons on diverse weather degradation datasets. From top to bottom: rain and haze removal (Outdoor-Rain), raindrop removal (RainDrop), and snow removal (Snow100K). CPL removes adverse weather artifacts while preserving fine details.}

    \label{fig_allweather}
\end{figure*}

\begin{table*}[h!]
\centering
\footnotesize
\renewcommand{\arraystretch}{1.3}
\setlength{\tabcolsep}{2pt}
\caption{Quantitative comparisons on Real-World Deweathering. We report PSNR, SSIM, and LPIPS for all tasks.}
\label{tab:weatherbench_results}
\resizebox{\linewidth}{!}{
\begin{tabular}{c|l|l|ccc|ccc|ccc|ccc}
\toprule
\multirow{2}{*}{\textbf{Type}} & \multirow{2}{*}{\textbf{Method}} & \multirow{2}{*}{\textbf{Venue}} & \multicolumn{3}{c|}{\textbf{Dehaze}} & \multicolumn{3}{c|}{\textbf{Derain}} & \multicolumn{3}{c|}{\textbf{Desnow}} & \multicolumn{3}{c}{\textbf{Average}} \\ \cline{4-15}
 & & & \textbf{PSNR}$\uparrow$ & \textbf{SSIM}$\uparrow$ & \textbf{LPIPS}$\downarrow$ & \textbf{PSNR}$\uparrow$ & \textbf{SSIM}$\uparrow$ & \textbf{LPIPS}$\downarrow$ & \textbf{PSNR}$\uparrow$ & \textbf{SSIM}$\uparrow$ & \textbf{LPIPS}$\downarrow$ & \textbf{PSNR}$\uparrow$ & \textbf{SSIM}$\uparrow$ & \textbf{LPIPS}$\downarrow$ \\
\hline
\multirow{7}{*}{\makecell{\rotatebox{90}{\textit{General}}}} & DehazeFormer~\cite{DehazeFormer} & TIP'23 & 24.12 & 0.745 & 0.345 & 36.05 & 0.954 & 0.181 & 28.88 & 0.849 & 0.178 & 29.68 & 0.849 & 0.235 \\
 & DCMPNet~\cite{Zhang2024DCMPNet} & CVPR'24 & 21.18 & 0.506 & 0.491 & 32.04 & 0.876 & 0.282 & 24.81 & 0.614 & 0.546 & 26.01 & 0.665 & 0.440 \\
 & DRSformer~\cite{chen2023drsformer} & CVPR'23 & 19.95 & 0.694 & 0.404 & 33.98 & 0.943 & 0.209 & 28.00 & 0.836 & 0.197 & 27.31 & 0.824 & 0.270 \\
 & NeRD-Rain~\cite{Chen2024nerd-rain} & CVPR'24 & 21.52 & 0.718 & 0.386 & 35.74 & 0.950 & 0.182 & 28.87 & 0.851 & 0.182 & 28.71 & 0.840 & 0.250 \\
 & SnowFormer~\cite{Chen2022Snowformer} & arXiv'22 & 22.71 & 0.736 & \underline{0.305} & 35.18 & 0.951 & 0.155 & 29.30 & 0.868 & \underline{0.143} & 29.06 & 0.852 & \underline{0.201} \\
 & MPRNet~\cite{Zamir_2021_CVPR_mprnet} & CVPR'21 & 23.27 & 0.739 & 0.355 & 36.14 & 0.954 & 0.171 & 29.18 & 0.860 & 0.177 & 29.53 & 0.851 & 0.234 \\
 & Restormer~\cite{zamir2022restormer} & CVPR'22 & 19.30 & 0.687 & 0.412 & 34.49 & 0.945 & 0.197 & 27.95 & 0.836 & 0.197 & 27.25 & 0.823 & 0.269 \\ \hline
\multirow{9}{*}{\makecell{\rotatebox{90}{\textit{All-in-One}}}} & AirNet & CVPR'22 & 20.94 & 0.705 & 0.383 & 33.59 & 0.942 & 0.224 & 22.06 & 0.780 & 0.291 & 25.53 & 0.809 & 0.299 \\
 & TransWeather~\cite{Transweather} & CVPR'22 & 19.79 & 0.680 & 0.397 & 29.34 & 0.903 & 0.294 & 24.96 & 0.796 & 0.231 & 24.70 & 0.793 & 0.307 \\
 & WGWS-Net~\cite{weather_data} & CVPR'23 & 13.79 & 0.603 & 0.535 & \textbf{37.08} & \textbf{0.961} & \textbf{0.117} & 20.81 & 0.780 & 0.248 & 23.89 & 0.781 & 0.300 \\
 & DiffUIR~\cite{Zheng2024DiffUIR} & CVPR'24 & 22.74 & 0.744 & 0.355 & 35.93 & 0.955 & 0.172 & 29.50 & \underline{0.870} & 0.162 & 29.39 & \underline{0.856} & 0.230 \\
 & MWFormer~\cite{Zhu2024MWFormer} & TIP'24 & \underline{24.42} & \underline{0.746} & \textbf{0.284} & 35.15 & 0.951 & \underline{0.153} & \underline{29.98} & \textbf{0.872} & \textbf{0.133} & \underline{29.85} & \underline{0.856} & \textbf{0.190} \\
 & Histoformer~\cite{Histoformer_SunRGWC24} & ECCV'24 & 17.69 & 0.669 & 0.437 & 30.70 & 0.916 & 0.279 & 25.39 & 0.808 & 0.225 & 24.59 & 0.798 & 0.314 \\
 & AdaIR~\cite{cui2025adair} & ICLR'25 & 23.08 & 0.731 & 0.351 & 34.87 & 0.946 & 0.192 & 28.44 & 0.837 & 0.179 & 28.80 & 0.838 & 0.240 \\
 \cline{2-15}
 & PromptIR~\cite{Potlapalli2023promptir} & NeurIPS'23 & 21.11 & 0.713 & 0.375 & 34.54 & 0.944 & 0.198 & 27.93 & 0.836 & 0.195 & 27.86 & 0.831 & 0.256 \\
 & \cellcolor{my_color}\textbf{PromptIR+CPL} & \cellcolor{my_color}2025 &
\cellcolor{my_color}\textbf{24.48} & \cellcolor{my_color}\textbf{0.747} & \cellcolor{my_color}0.339 &
\cellcolor{my_color}\underline{36.57} & \cellcolor{my_color}\underline{0.958} & \cellcolor{my_color}0.168 &
\cellcolor{my_color}\textbf{30.02} & \cellcolor{my_color}0.863 &
\cellcolor{my_color}0.172 &
\cellcolor{my_color}\textbf{30.35} & \cellcolor{my_color}\textbf{0.856} & \cellcolor{my_color}0.226 \\
\bottomrule
\end{tabular}
}
\end{table*}
\begin{figure*}[t]
    \centering
    \includegraphics[width=\textwidth]{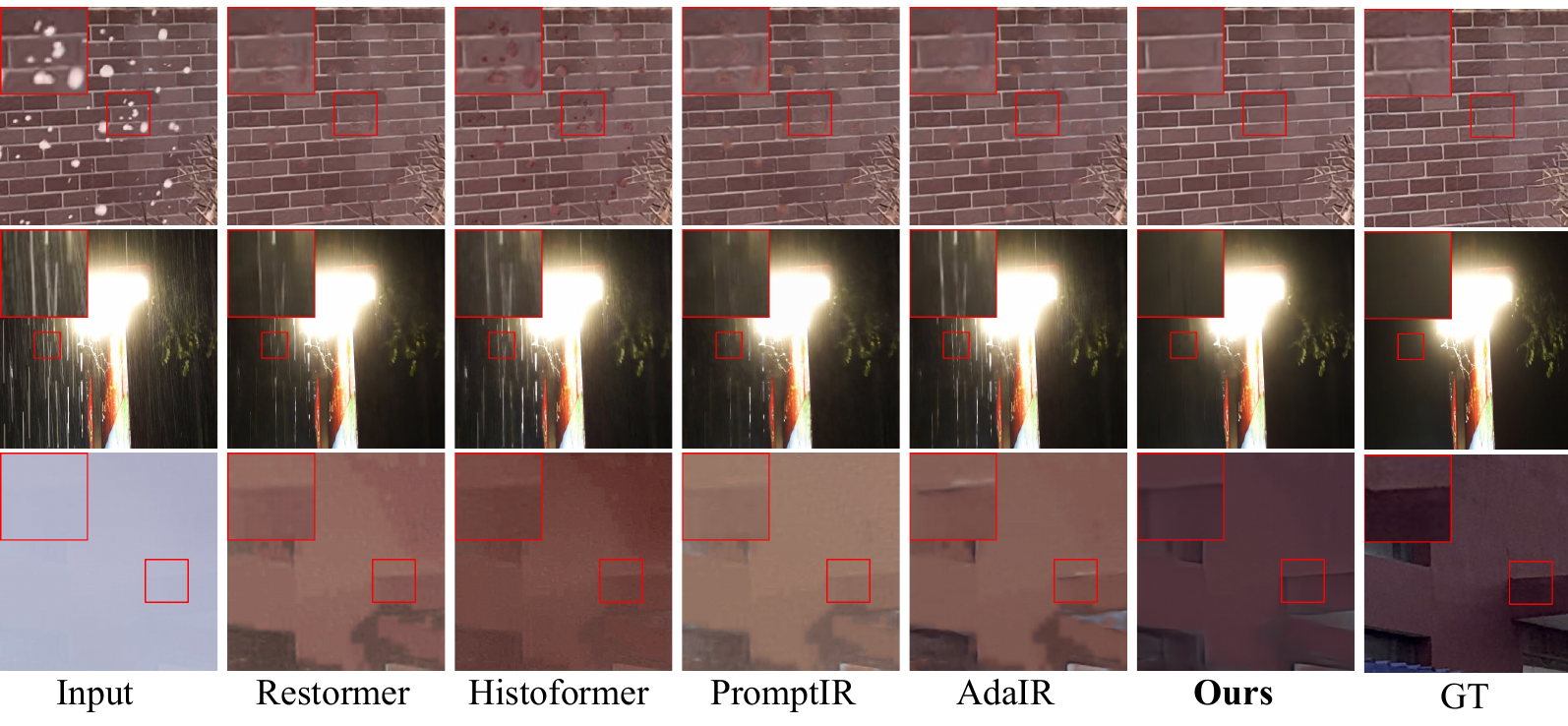}
\caption{Visual comparisons on real-world adverse weather scenes from WeatherBench \cite{weatherbench_data}. From top to bottom: snow, night rain, and haze. CPL improves visibility and texture details compared with baseline methods.}

    \label{fig_weather_bench}
\end{figure*}

\begin{table*}[tb]
\setlength{\abovecaptionskip}{0pt}
\caption{Quantitative evaluation on the composite degradation benchmark following \cite{guo2024onerestore}. Single and multiple composite degradations are evaluated on the CDD-11 dataset.}

  \label{tab_composite}
  \centering
  \large
\renewcommand\arraystretch{1.2}
\small
\resizebox{\linewidth}{!}{
\begin{tabular}{l|l|cccc|ccccc|cc|c}
\toprule[1pt]
\textbf{Method}& \textbf{Venue}& \textbf{l} & \textbf{h} & \textbf{r}& \textbf{s}& \textbf{l+h} & \textbf{l+r} & \textbf{l+s}& \textbf{h+r} & \textbf{h+s} & \textbf{l+h+r} & \textbf{l+h+s}  & \textbf{Avg.}\\
\hline
AirNet\cite{Li22AirNet} & CVPR'22 & 24.83 & 24.21 & 26.55 & 26.79 & 23.23 & 22.82 & 23.29 & 22.21 & 23.29 & 21.80 & 22.24 & 23.75  \\
TransWeather\cite{Transweather}&CVPR'22 & 23.39 & 23.95 & 26.69 & 25.74 & 22.24 & 22.62 & 21.80 & 23.10 & 22.34 & 21.55 & 21.01 & 23.13 \\
WeatherDiff\cite{ozan2023weatherdiff} &TPAMI'23 & 23.58 & 21.99 & 24.85 & 24.80 & 21.83 & 22.69 & 22.12 & 21.25 & 21.99 & 21.23 & 21.04  & 22.49 \\
WGWSNet\cite{weather_data} & CVPR'23& 24.39 & 27.90 & 33.15 & 34.43& 24.27& 25.06& 24.60& 27.23& 27.65& 23.90 & 23.97 & 26.96 \\
InstructIR\cite{Conde2024InstructIR}&ECCV'24 & \underline{26.70} & 32.61 & \underline{33.51} & 34.45 & 24.36 & 25.41 & \underline{25.63} & 28.80 & 29.64 & 24.84 & 24.32 & 28.21 \\

OneRestore\cite{guo2024onerestore}&ECCV'24& 26.55 & \underline{32.71} & 33.48 & \underline{34.50} & \underline{26.15} & \underline{25.83} & 25.56 & \underline{30.27} & \textbf{30.46} & \underline{25.18} & \underline{25.28}  & \underline{28.47}\\
 \hline
PromptIR\cite{Potlapalli2023promptir} & NeurIPS’23 & 26.32 & 26.10 & 31.56 & 31.53 & 24.49 & 25.05 & 24.51 & 24.54 & 23.70 & 23.74 & 23.33 & 25.90\\
\cellcolor{my_color}\textbf{PromptIR+CPL} 
&\cellcolor{my_color} 2025
& \cellcolor{my_color}\textbf{27.35} 
& \cellcolor{my_color}\textbf{32.79} 
& \cellcolor{my_color}\textbf{34.60} 
& \cellcolor{my_color}\textbf{35.35} 
& \cellcolor{my_color}\textbf{26.18} 
& \cellcolor{my_color}\textbf{26.32} 
& \cellcolor{my_color}\textbf{26.10}
& \cellcolor{my_color}\textbf{30.40} 
& \cellcolor{my_color}\underline{29.95} 
& \cellcolor{my_color}\textbf{25.40} 
& \cellcolor{my_color}\textbf{25.35} 
& \cellcolor{my_color}\textbf{29.07}\\
\bottomrule[1pt]
\hline
\end{tabular}
}

\end{table*}
\begin{figure*}[!h]
    \centering
    \includegraphics[width=\textwidth]{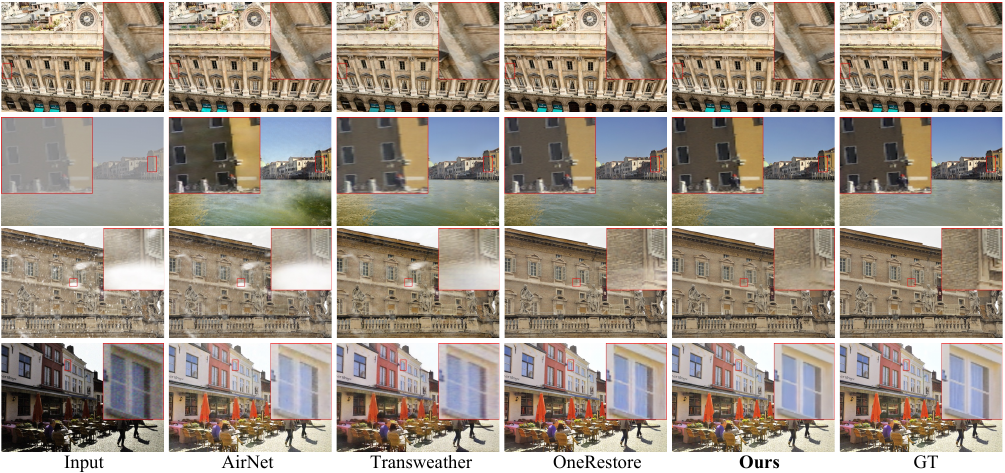}
    \vspace{-5mm}
\caption{Visual comparisons on single-degradation cases from the composite degradation dataset. From top to bottom: deraining, dehazing, desnowing, and low-light enhancement. CPL removes degradations while maintaining scene structures and natural color appearance.}

    \label{fig_cdd_single}
\end{figure*}
\begin{figure*}[t]
    \centering
    \includegraphics[width=\textwidth]{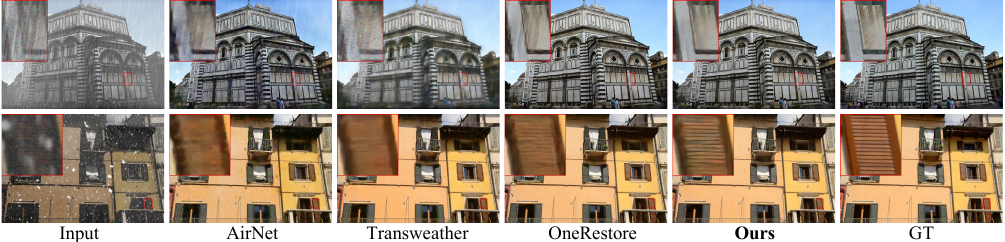}
\caption{Visual comparisons on composite degradations from the composite degradation dataset: low-light+haze+rain (top row) and low-light+haze+snow (bottom row). CPL simultaneously reduces multiple degradations and preserves structural details and color fidelity.}

    \label{fig_cdd_third}
    
\end{figure*}

\begin{figure}[t]
    \centering
    \includegraphics[width=\linewidth]{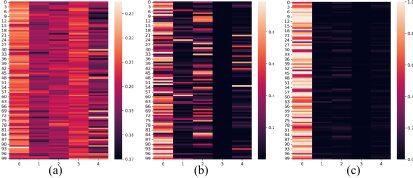}
\caption{Visualization of prompt selection patterns across 100 test images from the Rain100L dataset, where the x-axis represents prompt indices and the y-axis represents sample indices. (a) Baseline PromptIR with adaptive prompts. (b) CPL with the SPM. (c) CPL with both SPM and CPR.}
    \label{fig_moe_prompt}
\end{figure}

\paragraph{Seven-Task Comprehensive Evaluation}
We conduct experiments on a comprehensive seven-task benchmark \cite{MioIR}. As shown in Table \ref{table_7_task}, our approach achieves the highest average PSNR over all seven tasks, improving upon PromptIR-S by 0.68 dB and upon the PromptIR baseline by 1.42 dB. These performance gains are consistent across all degradation types. When compared to other training strategies like sequential learning (S) and explicit prompt learning (EP), CPL consistently delivers superior results. For instance, while Restormer+EP performs well on low-light enhancement, it underperforms on other tasks. In contrast, PromptIR+CPL maintains robust performance across the full spectrum of degradations without compromising specific tasks, validating its effectiveness in mitigating task interference.

\paragraph{All-Weather Restoration Experiment}
We evaluate our method on all-weather degradation scenarios \cite{Transweather}, including snow, rain, and raindrops. As shown in Table \ref{tab_allweather}, our approach achieves the best overall average performance, representing a 0.94 dB improvement over the original PromptIR and surpassing Histoformer \cite{Histoformer_SunRGWC24} by 0.06 dB. Notably, our method demonstrates consistent gains across all datasets: 0.94 dB on Snow100K-S, 0.93 dB on Snow100K-L, 1.36 dB on Outdoor-Rain, and 0.53 dB on RainDrop. This balanced improvement supports CPL's effectiveness in enhancing task-specific adaptation without compromising generalization capabilities. Figure \ref{fig_allweather} presents visual comparisons across different weather degradation types. For snow removal (first column), PromptIR+CPL completely eliminates snow particles while preserving intricate background details and natural colors. Other methods either leave snow residuals (TransWeather, WeatherDiff) or produce color distortions and over-smoothing (original PromptIR).

\paragraph{Real-World All-Weather Restoration Experiment}
To validate the generalization capability of CPL beyond synthetic data, we conduct a comprehensive evaluation on the real-world WeatherBench dataset \cite{weatherbench_data}, as detailed in \Cref{tab:weatherbench_results}. Quantitatively, our model attains an average PSNR of 30.35 dB, which is 2.49 dB higher than the PromptIR baseline. The superior LPIPS scores further indicate that these gains translate into perceptible improvements in image quality. As illustrated in Figure \ref{fig_weather_bench}, baseline methods often struggle with the complexity of real-world degradations. For instance, in the snow removal example, competing methods fail to distinguish between white particle artifacts and the underlying texture, leading to residual noise. In contrast, CPL effectively suppresses weather artifacts while maintaining structural fidelity and color accuracy.

\paragraph{Composite Degradation Experiment}
To evaluate the capability of handling coexisting degradations, we use the CDD-11 dataset \cite{guo2024onerestore}, which contains both single and composite degradations. As shown in Table~\ref{tab_composite}, PromptIR+CPL achieves the highest average PSNR and ranks first in 10 out of 11 categories, with an average of 29.07 dB (0.60 dB higher than OneRestore \cite{guo2024onerestore}). These results suggest that CPL improves robustness in composite degradation scenarios. For single degradations, PromptIR+CPL yields large gains over PromptIR, such as 6.69 dB in dehazing and 3.82 dB in desnowing. The advantage is maintained as the degradation complexity increases. For mixed degradations, we observe clear improvements in challenging combinations such as haze+rain (30.40 dB), and even for third-order composite degradations (e.g., low-light+haze+rain), where the gain over the baseline reaches 1.66 dB. Figure~\ref{fig_cdd_single} presents visual comparisons for single degradations; for low-light enhancement, CPL provides more balanced illumination while preserving color tones and texture details. Figure~\ref{fig_cdd_third} shows examples of composite degradations, where CPL reduces multiple artifacts simultaneously and better preserves structural details and color consistency.

\subsection{Ablation Studies}
\label{subsec:ablation}
{In this section, we conduct a series of detailed ablation experiments to dissect the key components and hyperparameters of our CPL framework. We analyze the impact of prompt sparsity in SPM, the generalizability of CPL across diverse architectures, and the sensitivity to the CPR loss weight.}

\paragraph{Effect of Prompt Sparsity in SPM} 
To rigorously isolate the contribution of the SPM, we conduct this ablation \textit{without} the proposed CPR. We investigate the impact of the sparsity level by varying the number of activated prompts, $k$, in the top-$k$ gating mechanism. As shown in \Cref{tab:ablation_k_3task,tab:ablation_k_5task}, the optimal choice of $k$ reveals a trade-off between specialization and collaboration. On the three-task benchmark, where degradations share certain statistical similarities (e.g., rain streaks and noise), setting $k=2$ yields the highest PSNR (32.54 dB). This suggests that allowing a slight relaxation of sparsity, enabling two complementary prompts to collaborate, provides flexibility for handling overlapping degradation patterns. However, increasing $k$ further to three results in a performance drop, confirming that excessive prompt activation reintroduces the representation redundancy that SPM is designed to mitigate. Conversely, on the more diverse five-task benchmark, which involves distinct physical mechanisms (e.g., global illumination in low-light vs. local motion blur), resulting in $k=1$ proves optimal. This constraint compels the model to rely on a single, highly specialized prompt representation, effectively preventing interference from irrelevant task priors. 

Based on these findings, we adopt $k=1$ as the default setting for all main experiments to prioritize clear task boundaries in complex scenarios. {It is worth noting that} the results in this ablation are lower than the final performance reported in \Cref{tab_three_task,tab_five_task}. This gap showcases the significant contribution of CPR, which further boosts performance by explicitly enforcing functional alignment between the selected prompt and the restoration objective.

\paragraph{Prompt Selection Patterns}
To further examine the effect of the proposed modules, we visualize prompt selection patterns during inference on the Rain100L test set in Figure~\ref{fig_moe_prompt}. The baseline PromptIR model (Figure~\ref{fig_moe_prompt}(a)) shows a diffuse selection distribution, where activations are spread across multiple prompt indices for most samples. This observation is consistent with our entropy analysis and suggests that adaptive prompts without additional constraints tend to develop redundant and overlapping representations. After introducing the SPM (Figure~\ref{fig_moe_prompt}(b)), the selection becomes more concentrated, with most samples relying on a small subset of prompts, indicating more task-specific usage of prompt capacity. When CPR is further added (Figure~\ref{fig_moe_prompt}(c)), the selection pattern becomes even more consistent across samples. This progression from (a) to (c) is in line with our design goal: combining sparse selection with contrastive regularization encourages prompts that are more specialized and better aligned with the restoration tasks.

\paragraph{Prompt Scalability Analysis}
To evaluate how effectively our sparse prompt design scales compared with dense alternatives, we increase the number of available prompts from 5 to 15 for both PromptIR and CPL on the CDD-11 benchmark. Two observations emerge from \Cref{tab:scalability_efficiency}. First, the dense ensembling strategy in PromptIR exhibits clear diminishing returns: tripling the number of prompts and parameters yields only a marginal average PSNR gain of +0.20 dB (from 25.90 dB to 26.10 dB), together with an increase in FLOPs. This suggests that simply adding more dense prompts largely amplifies existing redundancy rather than providing genuinely new task-specialized capacity. In contrast, CPL shows a more favorable scaling behavior. Under the same increase in prompt capacity, our model improves by +0.78 dB (from 29.07 dB to 29.85 dB), which is almost four times the gain observed for PromptIR, while keeping the inference FLOPs nearly constant due to the top-$k$ sparse routing in SPM. This pattern is consistent with our hypothesis that unconstrained dense ensembling tends to suffer from representation redundancy and limited capacity utilization, whereas enforcing sparsity at the prompt level allows the model to better exploit a larger and more diverse prompt pool for performance scaling.

\begin{table}[h!]
\caption{{Ablation study on the value of $k$ in our SPM on the three-task benchmark. }}
\label{tab:ablation_k_3task}
\centering
\renewcommand\arraystretch{1.2}

\resizebox{\linewidth}{!}{
\begin{tabular}{l|ccccc|c}
\toprule[1pt]
\multirow{2}{*}{\textbf{Method}} & \textbf{Dehazing} & \textbf{Deraining} & \multicolumn{3}{c|}{\textbf{Denoising on BSD68}} & \multirow{2}{*}{\textbf{Average}} \\
& on SOTS & on Rain100L & $\sigma$=15 & $\sigma$=25 & $\sigma$=50 & \\
\hline
Baseline & 30.58 / 0.974 & 36.37 / 0.972 & 33.98 / 0.933 & 31.31 / 0.888 & 28.06 / 0.799 & 32.06 / 0.913 \\
Ours ($k=1$) & 30.60 / 0.975 & 38.30 / 0.983 & 34.00 / 0.930 & 31.35 / 0.884 & 28.15 / 0.795 & 32.48 / 0.913 \\
\textbf{Ours ($k=2$)} & \textbf{30.66 / 0.977} & \textbf{38.45 / 0.984} & \textbf{34.05 / 0.931} & \textbf{31.40 / 0.885} & \textbf{28.15 / 0.795} & \textbf{32.54 / 0.914} \\
Ours ($k=3$) & 30.41 / 0.972 & 37.57 / 0.979 & 33.93 / 0.928 & 31.26 / 0.882 & 27.98 / 0.792 & 32.23 / 0.911 \\
\bottomrule[1pt]
\end{tabular}
}
\end{table}

\begin{table}[h!]
\caption{{Ablation study on the value of $k$ in our SPM on the five-task benchmark. }}
\label{tab:ablation_k_5task}
\centering
\renewcommand\arraystretch{1.2}

\resizebox{\linewidth}{!}{
\begin{tabular}{l|ccccc|c}
\toprule[1pt]
\multirow{2}{*}{\bf Method} & {\bf Denoising} & {\bf Dehazing} & {\bf Deraining} & {\bf Deblurring} & {\bf Low-light} & \multirow{2}{*}{\bf Average} \\
& CBSD68 & SOTS & Rain100L & GoPro & LOL & \\
\hline
Baseline & 31.47 / 0.886 & 26.54 / 0.949 & 36.37 / 0.970 & 28.71 / 0.881 & 22.68 / 0.832 & 29.15 / 0.904 \\
\textbf{Ours ($k=1$)} & \textbf{31.35 / 0.886} & \textbf{29.67 / 0.963} & \textbf{37.47 / 0.974} & \textbf{28.67 / 0.881} & \textbf{23.12 / 0.844} & \textbf{30.06 / 0.908} \\
\textbf{Ours ($k=2$)} & 31.29 / 0.885 & 28.83 / 0.959 & 36.91 / 0.972 & 28.44 / 0.885 & 22.97 / 0.838 & 29.68 / 0.906 \\
\textbf{Ours ($k=3$)} & 31.28 / 0.885 & 28.37 / 0.952 & 36.75 / 0.972 & 28.42 / 0.885 & 22.92 / 0.837 & 29.54 / 0.905 \\
\bottomrule[1pt]
\end{tabular}
}\vspace{-5mm}
\end{table}

\paragraph{Computational Efficiency Analysis}
The same experiment also highlights the computational efficiency of our framework. As detailed in \Cref{tab:scalability_efficiency}, even in a parameter-matched setting with 5 prompts, our SPM is inherently more efficient, reducing FLOPs by 5\% compared to PromptIR while delivering significantly higher accuracy. More critically, the efficiency gap widens as the model scales. Because the inference cost of SPM depends only on the number of activated experts ($k=1$), our model's FLOPs remain constant at 43.18 GFLOPs regardless of the total number of prompts available. Conversely, PromptIR's computational cost increases linearly with the number of prompts. Consequently, our 15-prompt model is better than its 15-prompt PromptIR counterpart and {14.2\% more computationally efficient}. This demonstrates that our CPL framework achieves superior performance and scalability without the typical trade-off in computational cost, offering a more practical solution for real-world deployment.

\begin{table*}[h!]
\setlength{\abovecaptionskip}{5pt}
\caption{{Efficiency and scalability comparison on the CDD-11 benchmark. Our sparse prompt design is more efficient and scales far more effectively than the dense approach of PromptIR.}}
\label{tab:scalability_efficiency}
\centering
\renewcommand{\arraystretch}{1.3}
\small
\resizebox{\linewidth}{!}{
\begin{tabular}{c|c|cc|ccccccccccc|c}
\toprule[1pt]
\textbf{Method} & \textbf{Prompts} & \textbf{Params. (M)} & \textbf{FLOPs (G)} & \textbf{l} & \textbf{h} & \textbf{r} & \textbf{s} & \textbf{l+h} & \textbf{l+r} & \textbf{l+s} & \textbf{h+r} & \textbf{h+s} & \textbf{l+h+r} & \textbf{l+h+s} & \textbf{Avg.} \\
\hline
\multirow{3}{*}{PromptIR  } & 5 & 35.38 & 45.55 & 26.32 & 26.10 & 31.56 & 31.53 & 24.49 & 25.05 & 24.51 & 24.54 & 23.70 & 23.74 & 23.33 & 25.90 \\
& 10 & 37.76 & 47.93 & 26.41 & 26.19 & 31.65 & 31.60 & 24.62 & 25.15 & 24.60 & 24.68 & 23.81 & 23.85 & 23.45 & 26.05 \\
& 15 & 40.14 & 50.31 & 26.45 & 26.23 & 31.69 & 31.62 & 24.68 & 25.20 & 24.63 & 24.75 & 23.88 & 23.90 & 23.51 & 26.10 \\
\hline
\multirow{3}{*}{Ours} & 5 & 35.38 & 43.18  & 27.35 & 32.79 & 34.60 & 35.35 & 26.18 & 26.32 & 26.10 & 30.40 & 29.95 & 25.40 & 25.35 & 29.07 \\
& 10 & 37.76 & 43.18   & 27.48 & 34.91 & 35.00 & 35.96 & 26.70 & 26.55 & 26.38 & 31.52 & 31.49 & 25.79 & 25.83 & 29.60 \\
& 15 & 40.14 & 43.18  & 27.56 & 35.85 & 35.23 & 36.25 & 26.95 & 26.65 & 26.54 & 32.14 & 32.07 & 26.07 & 26.01 & 29.85 \\
\bottomrule[1pt]
\end{tabular}
}
\end{table*}
 
\paragraph{Cross-Architecture Evaluation}
To validate the plug-and-play nature and architecture-agnostic benefits of our CPL framework, we integrated it into four diverse, high-performing restoration backbones: NAFNet and FSNet (CNN-based), DRSformer (Transformer-based), and MambaIR (state-space model). We retrained each model on the five-task benchmark, and the results are presented in \Cref{tab:plug_and_play_validation}. Across all backbones, CPL yields consistent improvements in average PSNR, with gains ranging from +1.01 dB for MambaIR to +1.15 dB for FSNet. The presence of similar gains across architectures with markedly different inductive biases suggests that CPL acts as a generally applicable enhancement module, rather than a mechanism tailored to a specific network design.

\begin{table*}[h!]
\setlength{\abovecaptionskip}{5pt}
\caption{Plug-and-play evaluation of CPL integrated into different restoration backbones on the five-task benchmark.}

\label{tab:plug_and_play_validation}
\renewcommand{\arraystretch}{1.3}

\centering
\resizebox{0.9\linewidth}{!}{
\begin{tabular}{l|l|ccccc|c} \toprule[1pt] 
\multirow{2}{*}{\bf Method} & \multirow{2}{*}{\bf Venue} & \multicolumn{1}{c}{\bf Denoising} & \multicolumn{1}{c}{\bf Dehazing} & \multicolumn{1}{c}{\bf Deraining} & \multicolumn{1}{c}{\bf Deblurring} & \multicolumn{1}{c}{\bf Low-light} & \multirow{2}{*}{\bf Average} \\

& & \multicolumn{1}{c}{CBSD68} & \multicolumn{1}{c}{SOTS} & \multicolumn{1}{c}{Rain100L} & \multicolumn{1}{c}{GoPro} & \multicolumn{1}{c}{LOL} & \\
\hline NAFNet~\cite{chen2022simple} & \multirow{2}{*}{ECCV'22} & 31.02 / 0.883 & 25.23 / 0.939 & 35.56 / 0.967 & 26.53 / 0.808 & 20.49 / 0.809 & 27.76 / 0.881 \\ 
\textbf{NAFNet+CPL} & & \cellcolor{my_color}31.15 / 0.883 & \cellcolor{my_color}28.18 / 0.951 & \cellcolor{my_color}37.47 / 0.973 & \cellcolor{my_color}28.78 / 0.829 & \cellcolor{my_color}22.27 / 0.827 & \cellcolor{my_color}\textbf{29.57 / 0.893} \\
\hline DRSformer~\cite{chen2023drsformer} & \multirow{2}{*}{CVPR'23} & 30.97 / 0.881 & 24.66 / 0.931 & 33.45 / 0.953 & 25.56 / 0.780 & 21.77 / 0.821 & 27.28 / 0.873 \\ 
\textbf{DRSformer+CPL} & & \cellcolor{my_color}31.18 / 0.881 & \cellcolor{my_color}27.19 / 0.945 & \cellcolor{my_color}35.20 / 0.961 & \cellcolor{my_color}27.61 / 0.801 & \cellcolor{my_color}22.42 / 0.833 & \cellcolor{my_color}\textbf{28.72 / 0.884} \\
\hline FSNet~\cite{cui2024FSNet} & \multirow{2}{*}{TPAMI'23} & 31.33 / 0.883 & 25.53 / 0.943 & 36.07 / 0.968 & 28.32 / 0.869 & 22.29 / 0.829 & 28.71 / 0.898 \\
\textbf{FSNet+CPL} & & \cellcolor{my_color}31.44 / 0.884 & \cellcolor{my_color}28.71 / 0.954 & \cellcolor{my_color}37.82 / 0.975 & \cellcolor{my_color}28.73 / 0.877 & \cellcolor{my_color}22.62 / 0.834 & \cellcolor{my_color}\textbf{29.86 / 0.905} \\
\hline MambaIR~\cite{guo2024mambair} & \multirow{2}{*}{ECCV'24} & 31.41 / 0.884 & 25.81 / 0.944 & 36.55 / 0.971 & 28.61 / 0.875 & 22.49 / 0.832 & 28.97 / 0.901 \\
\textbf{MambaIR+CPL} & & \cellcolor{my_color}31.50 / 0.886 & \cellcolor{my_color}28.84 / 0.958 & \cellcolor{my_color}37.94 / 0.976 & \cellcolor{my_color}28.83 / 0.884 & \cellcolor{my_color}22.79 / 0.836 & \cellcolor{my_color}\textbf{29.98 / 0.908} \\
\bottomrule[1pt] 
\end{tabular} 
}\vspace{-5mm}
\end{table*}
\paragraph{Sensitivity Analysis on the CPR Loss Weight}
We analyze the model's sensitivity to the hyperparameter $\alpha$ in Eq.~\eqref{eq:total_loss}, which balances the contrastive prompt regularization $\mathcal{L}_{\mathrm{cpr}}$ against the primary reconstruction loss. We trained models on both the three-task and five-task benchmarks while varying $\alpha$ across a wide range from 0.001 to 0.3. As shown in \Cref{tab:ablation_alpha}, our method exhibits strong robustness to this choice. Performance remains stable across the range of $10^{-3}$ to $10^{-2}$, with optimal results achieved at $\alpha=0.003$ for the three-task setting and $\alpha=0.01$ for the five-task setting. This indicates that a small weighting for the CPR term is sufficient to guide the model towards functional alignment without disrupting the primary restoration objective. We select $\alpha=0.01$ as a unified default for all experiments, as it provides near-optimal performance across benchmarks without requiring task-specific tuning. This stability underscores the reliability and ease of implementation of our CPR strategy.

\begin{table}[h!]
\caption{Sensitivity analysis of the CPR loss weight $\alpha$. Results are reported on both three-task and five-task benchmarks.}
\label{tab:ablation_alpha}
\centering
\renewcommand{\arraystretch}{1.5}
\resizebox{\linewidth}{!}{

\begin{tabular}{l|cccccc}
\toprule[1pt]
\textbf{Task} & $\alpha$ = 0.001 & $\alpha$ = 0.003 & $\alpha$ = 0.01 & $\alpha$ = 0.03 & $\alpha$ = 0.1 & $\alpha$ = 0.3 \\
\hline
Three-Task Benchmark & 32.78 & \textbf{32.81} & 32.78 & 32.75 & 32.75 & 32.69 \\
Five-Task Benchmark & 30.42 & 30.47 & \textbf{30.55} & 30.53 & 30.49 & 30.38 \\
\bottomrule[1pt]
\end{tabular}
}
\end{table}

\begin{table}[t]
\centering
\caption{Ablation study on the number of negative samples in CPR. Performance is evaluated on the three-task benchmark.}
\label{tab:neg_samples}
\begin{tabular}{lc}
\toprule
\textbf{Method} & \textbf{PSNR (dB)} \\
\midrule
Baseline (PromptIR) & 32.06 \\
CPR w/ 1 Negative Sample & 32.58 \\
CPR w/ 2 Negative Samples & 32.66 \\
CPR w/ 4 Negative Samples & \textbf{32.78} \\
CPR w/ 8 Negative Samples & 32.75 \\
\bottomrule
\end{tabular}\vspace{-5mm}
\end{table}

\paragraph{Effect of the Number of Negative Samples} 
We investigate the impact of the number of negative samples on restoration performance, as detailed in Table~\ref{tab:neg_samples}. The results reveal a clear relationship between the diversity of negative constraints and model efficacy. Introducing a single negative sample yields a substantial gain of 0.52 dB over the baseline, validating our premise that explicitly penalizing functional misalignment significantly enhances task specificity. As we increase the number to 4, the performance improves and peaks at 32.78 dB. This trend suggests that exposing the model to a wider variety of ``incorrect'' prompts helps establish sharper functional boundaries, compelling the network to adhere more strictly to the guidance of the target prompt. However, further increasing the number to 8 results in a slight saturation (32.75 dB), indicating diminishing returns and a potential dilution of the primary reconstruction gradient by redundant repulsive signals. We adopt 4 negative samples as the optimal trade-off between regularization strength and training stability.

\subsection{Limitations and Future Work}\label{subsec:discussion}
While our CPL framework demonstrates notable advancements in mitigating prompt redundancy and enhancing functional alignment, we acknowledge certain limitations. The main limitation lies in our CPR strategy. CPR currently constructs negative samples by treating all mismatched prompt-task pairings as equally negative. For instance, when restoring a low-light image, a prompt for a visually similar task like dehazing is penalized with the same repulsive force as a prompt for a highly dissimilar task like denoising. This uniform negative sampling strategy, while effective at enforcing task specificity, oversimplifies the complex web of inter-task relationships and does not explicitly leverage the semantic similarities and differences between various degradation types.

This limitation opens up several promising avenues for future research. A direct extension would be to develop more sophisticated negative sampling strategies for CPR. Instead of uniform sampling, one could implement a semantic-aware weighting scheme, where the repulsive force is modulated by the similarity between the ground-truth task and the negative prompt's task. This could allow the model to learn a more nuanced and structured prompt space. An even more ambitious direction would be to move beyond the binary contrastive paradigm altogether. Inspired by recent advancements in large model alignment, one could formulate the regularization as a preference learning problem. Rather than a simple positive/negative distinction, the model could be trained to learn a fine-grained preference for ``better'' restorations over ``worse'' ones, even when both are generated by sub-optimal prompts. Such a paradigm, conceptually related to Reinforcement Learning from Human Feedback (RLHF), could help the model learn a more detailed understanding of image quality and task relationships, further enhancing its generalization capabilities in complex, real-world scenarios.

\section{Conclusion}
\label{sec:conclusion}

{In this paper, we address the inherent challenges of representation redundancy and functional misalignment that hinder existing AiOIR frameworks. We propose a Contrastive Prompt Learning (CPL) framework that resolves this tension through two synergistic conceptual shifts in prompt design and regularization. First, the Sparse Prompt Module (SPM) tackles representation redundancy by leveraging principled sparsity to enhance the intrinsic quality and expressive power of each individual prompt. Complementing this, the Contrastive Prompt Regularization (CPR) mechanism addresses functional misalignment. It introduces a novel paradigm of functional regularization via a model-prompt decoupling strategy, which optimizes prompts for their behavioral alignment and reconstruction quality, rather than for abstract classification.} {Comprehensive experiments across a wide array of benchmarks validate the superiority of our approach. The consistent and significant performance improvements demonstrate that our synergistic design effectively addresses the core limitations of current prompt-based models. Beyond the immediate performance gains, our work suggests that an effective path to robust AiOIR is to co-design the intrinsic quality of prompts together with their functional alignment to the restoration model.}

\small
\bibliographystyle{IEEEtran}
\bibliography{main}

@String(TPAMI = {IEEE Trans. Pattern Anal. Mach. Intell.})

@String(IJCV = {Int. J. Comput. Vis.})

@String(CVPR = {IEEE Conf. Comput. Vis. Pattern Recog.})

@String(ICCV = {Int. Conf. Comput. Vis.})

@String(ECCV = {Eur. Conf. Comput. Vis.})

@String(NeurIPS = {Adv. Neural Inform. Process. Syst.})

@String(ICLR = {Int. Conf. Learn. Represent.})

@String(ICML = {Int. Conf. Mach. Learn.})

@String(IJCAI = {Int. Jt. Conf. Artif. Intell.})

@String(AAAI = {AAAI Conf. Artif. Intell.})

@String(ACMMM = {ACM Int. Conf. Multimedia})

@String(BMVC = {Brit. Mach. Vis. Conf.})

@String(ICIP = {IEEE Int. Conf. Image Process.})

@String(TIP = {IEEE Trans. Image Process.})

@String(TNNLS = {IEEE Trans. Neural Netw. Learn. Syst.})

@String(TITS = {IEEE Trans. Intell. Transport. Syst.})

@String(SPM = {IEEE Signal Process. Maga.})

@ARTICLE{Banham1997digitalsurvey,
  author={Banham, M.R. and Katsaggelos, A.K.},
  journal=SPM, 
  title={Digital image restoration}, 
  year={1997},
  volume={14},
  number={2},
  pages={24-41},
}

@inproceedings{weatherbench_data,
  author       = {Qiyuan Guan and
                  Qianfeng Yang and
                  Xiang Chen and
                  Tianyu Song and
                  Guiyue Jin and
                  Jiyu Jin},
  title        = {{WeatherBench}: {A} Real-World Benchmark Dataset for All-in-One Adverse Weather Image Restoration},
  booktitle      = ACMMM,
  pages = {12607–12613},
  year         = {2025}
}

@article{su2022survey,
	title = {A survey of deep learning approaches to image restoration},
	volume = {487},
	issn = {0925-2312},
	pages = {46--65},
	issue = {C},
	journal = {Neurocomput.},
	author = {Su, Jingwen and Xu, Boyan and Yin, Hujun},
	year = {2022-05},
	
}

@article{EladKV23DenoiseSurvey,
  author       = {Michael Elad and
                  Bahjat Kawar and
                  Gregory Vaksman},
  title        = {Image Denoising: The Deep Learning Revolution and Beyond - {A} Survey
                  Paper},
  journal      = {{SIAM} J. Imaging Sci.},
  volume       = {16},
  number       = {3},
  pages        = {1594--1654},
  year         = {2023}
}

@article{Zhang22DeblurSurvey,
  author       = {Kaihao Zhang and
                  Wenqi Ren and
                  Wenhan Luo and
                  Wei{-}Sheng Lai and
                  Bj{\"{o}}rn Stenger and
                  Ming{-}Hsuan Yang and
                  Hongdong Li},
  title        = {Deep Image Deblurring: {A} Survey},
  journal      = IJCV,
  volume       = {130},
  number       = {9},
  pages        = {2103--2130},
  year         = {2022}
}

@article{chen23DerainSurvey,
  author={Chen, Xiang and Pan, Jinshan and Dong, Jiangxin and Tang, Jinhui},
  journal=TPAMI, 
  title={Towards Unified Deep Image Deraining: A Survey and a New Benchmark}, 
  year={2025},
  volume={47},
  number={7},
  pages={5414-5433},
}

@article{Gui23DehazeSurvey,
  author       = {Jie Gui and
                  Xiaofeng Cong and
                  Yuan Cao and
                  Wenqi Ren and
                  Jun Zhang and
                  Jing Zhang and
                  Jiuxin Cao and
                  Dacheng Tao},
  title        = {A Comprehensive Survey and Taxonomy on Single Image Dehazing Based
                  on Deep Learning},
  journal      = {{ACM} Comput. Surv.},
  volume       = {55},
  number       = {13s},
  pages        = {279:1--279:37},
  year         = {2023}
}

@article{Li22LowlightSurvey,
  author       = {Chongyi Li and
                  Chunle Guo and
                  Linghao Han and
                  Jun Jiang and
                  Ming{-}Ming Cheng and
                  Jinwei Gu and
                  Chen Change Loy},
  title        = {Low-Light Image and Video Enhancement Using Deep Learning: {A} Survey},
  journal      = TPAMI,
  volume       = {44},
  number       = {12},
  pages        = {9396--9416},
  year         = {2022}
}

@article{jiang2024survey,
  author={Jiang, Junjun and Zuo, Zengyuan and Wu, Gang and Jiang, Kui and Liu, Xianming},
  journal=TPAMI, 
  title={A Survey on All-in-One Image Restoration: Taxonomy, Evaluation and Future Trends}, 
  year={2025},
  volume={47},
  number={12},
  pages={11892-11911},
}

@inproceedings{Potlapalli2023promptir,
  author       = {Vaishnav Potlapalli and
                  Syed Waqas Zamir and
                  Salman H. Khan and
                  Fahad Shahbaz Khan},
  title        = {{PromptIR}: Prompting for All-in-One Image Restoration},
  booktitle    =NeurIPS,
  year         = {2023}
}

@inproceedings{cui2025adair,
  author       = {Yuning Cui and
                  Syed Waqas Zamir and
                  Salman H. Khan and
                  Alois Knoll and
                  Mubarak Shah and
                  Fahad Shahbaz Khan},
  title        = {{AdaIR}: Adaptive All-in-One Image Restoration via Frequency Mining and Modulation},
  booktitle       =ICLR,
  year         = {2025}
}

@article{MioIR,
  author       = {Xiangtao Kong and
                  Chao Dong and
                  Lei Zhang},
  title        = {Towards Effective Multiple-in-One Image Restoration: {A} Sequential
                  and Prompt Learning Strategy},
  journal      = {CoRR},
  volume       = {abs/2401.03379},
  year         = {2024}
}

@inproceedings{Luo2024DA-CLIP,
  author       = {Ziwei Luo and
                  Fredrik K. Gustafsson and
                  Zheng Zhao and
                  Jens Sj{\"{o}}lund and
                  Thomas B. Sch{\"{o}}n},
  title        = {Controlling Vision-Language Models for Multi-Task Image Restoration},
booktitle       = ICLR,
  year         = {2024}
}

@inproceedings{hu2025promptir_cls,
  title={Universal Image Restoration Pre-training via Degradation Classification},
  author={Hu, JiaKui and Jin, Lujia and Yao, Zhengjian and Lu, Yanye},
booktitle       =ICLR,
  year={2025}
}

@inproceedings{li25mair,
  author       = {Boyun Li and Haiyu Zhao and Wenxin Wang and Peng Hu and
                   Yuanbiao Gou and
                  Xi Peng},
  title        = {{MaIR}: A Locality- and Continuity-Preserving Mamba for Image Restoration},
booktitle = CVPR,

  year         = {2025}
}

@Article{BSD500,
  author={Arbeláez, Pablo and Maire, Michael and Fowlkes, Charless and Malik, Jitendra},
  journal=TPAMI, 
  title={Contour Detection and Hierarchical Image Segmentation}, 
  year={2011},
  volume={33},
  number={5},
  pages={898-916},}

@article{WED_dataset,
	author    = {Ma, Kede and Duanmu, Zhengfang and Wu, Qingbo and Wang, Zhou and Yong, Hongwei and Li, Hongliang and Zhang, Lei}, 
	title     = {{Waterloo Exploration Database}: New Challenges for Image Quality Assessment Models}, 
	journal   =TIP,
	volume    = {26},
	number    = {2},
	pages     = {1004--1016},
	month	  = {Feb.},
	year      = {2017}
}

@inproceedings{chen2021IPT,
  author       = {Hanting Chen and
                  Yunhe Wang and
                  Tianyu Guo and
                  Chang Xu and
                  Yiping Deng and
                  Zhenhua Liu and
                  Siwei Ma and
                  Chunjing Xu and
                  Chao Xu and
                  Wen Gao},
  title        = {Pre-Trained Image Processing Transformer},
  booktitle    = CVPR,
  pages        = {12299--12310},
  year         = {2021},
}

@ARTICLE{huang2023memroyderain,
  author={Huang, Huaibo and Luo, Mandi and He, Ran},
  journal=TPAMI, 
  title={Memory Uncertainty Learning for Real-World Single Image Deraining}, 
  year={2023},
  volume={45},
  number={3},
  pages={3446-3460},
}

@ARTICLE{pan2018darkblur,
  author={Pan, Jinshan and Sun, Deqing and Pfister, Hanspeter and Yang, Ming-Hsuan},
  journal=TPAMI, 
  title={Deblurring Images via Dark Channel Prior}, 
  year={2018},
  volume={40},
  number={10},
  pages={2315-2328},
}

@inproceedings{Qin20FFANet,
  author       = {Xu Qin and
                  Zhilin Wang and
                  Yuanchao Bai and
                  Xiaodong Xie and
                  Huizhu Jia},
  title        = {{FFA-Net}: Feature Fusion Attention Network for Single Image Dehazing},
  booktitle    = AAAI,
  pages        = {11908--11915},
  year         = {2020}
}

@inproceedings{Li22AirNet,
  author       = {Boyun Li and
                  Xiao Liu and
                  Peng Hu and
                  Zhongqin Wu and
                  Jiancheng Lv and
                  Xi Peng},
  title        = {All-In-One Image Restoration for Unknown Corruption},
booktitle = CVPR,
  pages        = {17431--17441},
  year         = {2022}
}

@article{zhang2023practical,
  author       = {Kai Zhang and
                  Yawei Li and
                  Jingyun Liang and
                  Jiezhang Cao and
                  Yulun Zhang and
                  Hao Tang and
                  Deng{-}Ping Fan and
                  Radu Timofte and
                  Luc Van Gool},
  title        = {Practical Blind Image Denoising via Swin-Conv-UNet and Data Synthesis},
  journal      = {Mach. Intell. Res.},
  volume       = {20},
  number       = {6},
  pages        = {822--836},
  year         = {2023}
}

@inproceedings{zamir2022restormer,
	title = {Restormer: Efficient transformer for high-resolution image restoration},
	pages = {5718--5729},
	booktitle = CVPR,
	author = {Zamir, Syed Waqas and Arora, Aditya and Khan, Salman and Hayat, Munawar and Khan, Fahad Shahbaz and Yang, Ming-Hsuan},
	year = {2022},
}

@inproceedings{jiang2020multi,
	title = {Multi-scale progressive fusion network for single image deraining},
	pages = {8346--8355},
	booktitle = CVPR,
	author = {Jiang, Kui and Wang, Zhongyuan and Yi, Peng and Chen, Chen and Huang, Baojin and Luo, Yimin and Ma, Jiayi and Jiang, Junjun},
	year = {2020},
}

@inproceedings{cai2023retinexformer,
  author       = {Yuanhao Cai and
                  Hao Bian and
                  Jing Lin and
                  Haoqian Wang and
                  Radu Timofte and
                  Yulun Zhang},
  title        = {Retinexformer: One-stage Retinex-based Transformer for Low-light Image
                  Enhancement},
booktitle = ICCV,
  pages        = {12470--12479},
  year         = {2023}
}

@inproceedings{chen2022simple,
	title = {Simple baselines for image restoration},
	volume = {13667},
	pages = {17--33},
	booktitle = ECCV,
	author = {Chen, Liangyu and Chu, Xiaojie and Zhang, Xiangyu and Sun, Jian},
	year = {2022},
}

@inproceedings{Uformer,
	title = {Uformer: A general U-shaped transformer for image restoration},
	pages = {17662--17672},
	booktitle =CVPR,
	author = {Wang, Zhendong and Cun, Xiaodong and Bao, Jianmin and Zhou, Wengang and Liu, Jianzhuang and Li, Houqiang},
	year = {2022},
}

@article{wang2024gridformer,
  author       = {Tao Wang and
                  Kaihao Zhang and
                  Ziqian Shao and
                  Wenhan Luo and
                  Bj{\"{o}}rn Stenger and
                  Tong Lu and
                  Tae{-}Kyun Kim and
                  Wei Liu and
                  Hongdong Li},
  title        = {{GridFormer}: Residual Dense Transformer with Grid Structure for Image
                  Restoration in Adverse Weather Conditions},
  journal      = IJCV,
  volume       = {132},
  number       = {10},
  pages        = {4541--4563},
  year         = {2024}
}

@inproceedings{guo2024mambair,
  author       = {Hang Guo and
                  Jinmin Li and
                  Tao Dai and
                  Zhihao Ouyang and
                  Xudong Ren and
                  Shu{-}Tao Xia},
  title        = {MambaIR: {A} Simple Baseline for Image Restoration with State-Space
                  Model},
	booktitle = ECCV,
  volume       = {15076},
  pages        = {222--241},
  year         = {2024}
}

@InProceedings{zhang2023IDR,
  author    = {Jinghao Zhang and Jie Huang and Mingde Yao and Zizheng Yang and Hu Yu and Man Zhou and Feng Zhao},
  booktitle = CVPR,
  title     = {Ingredient-oriented Multi-Degradation Learning for Image Restoration},
  pages     = {5825--5835},
  year      = {2023},
}

@inproceedings{Radford2021CLIP,
  author       = {Alec Radford and
                  Jong Wook Kim and
                  Chris Hallacy and
                  Aditya Ramesh and
                  Gabriel Goh and
                  Sandhini Agarwal and
                  Girish Sastry and
                  Amanda Askell and
                  Pamela Mishkin and
                  Jack Clark and
                  Gretchen Krueger and
                  Ilya Sutskever},
  title        = {Learning Transferable Visual Models From Natural Language Supervision},
  booktitle    = ICML,
  volume       = {139},
  pages        = {8748--8763},
  year         = {2021}
}

@inproceedings{Lin2024Textual,
  author       = {Jingbo Lin and
                  Zhilu Zhang and
                  Yuxiang Wei and
                  Dongwei Ren and
                  Dongsheng Jiang and
                  Qi Tian and
                  Wangmeng Zuo},
  title        = {Improving Image Restoration Through Removing Degradations in Textual
                  Representations},
  booktitle = CVPR,
  pages        = {2866--2878},
  year         = {2024}
}

@inproceedings{wu2024harmony,
  author       = {Gang Wu and
                  Junjun Jiang and
                  Kui Jiang and
                  Xianming Liu},
  title        = {Harmony in Diversity: Improving All-in-One Image Restoration via Multi-Task
                  Collaboration},
  booktitle    = ACMMM,
  pages        = {6015--6023},
  year         = {2024}
}

@inproceedings{wu2025debiased,
  title={Debiased All-in-one Image Restoration with Task Uncertainty Regularization},
  author={Wu, Gang and Jiang, Junjun and Wang, Yijun  and Jiang,  Kui and  Liu, Xianming},
  booktitle=AAAI,
  year={2025},
  pages={8386--8394}
}

@inproceedings{qin2024maskedIR,
  author       = {Chu{-}Jie Qin and
                  Ruiqi Wu and
                  Zikun Liu and
                  Xin Lin and
                  Chun{-}Le Guo and
                  Hyun Hee Park and
                  Chongyi Li},
  title        = {Restore Anything with Masks: Leveraging Mask Image Modeling for Blind
                  All-in-One Image Restoration},
booktitle =ECCV,
  volume       = {15103},
  pages        = {364--380},
  year         = {2024}
}

@ARTICLE{Gui2024ContrastiveSurvey,
author={Gui, Jie and Chen, Tuo and Zhang, Jing and Cao, Qiong and Sun, Zhenan and Luo, Hao and Tao, Dacheng},
journal=TPAMI,
title={{ A Survey on Self-Supervised Learning: Algorithms, Applications, and Future Trends }},
year={2024},
volume={46},
number={12},
ISSN={1939-3539},
pages={9052-9071},
}

@inproceedings{wu2021contrastive1,
	title = {Contrastive learning for compact single image dehazing},
	pages = {10551--10560},
	booktitle =CVPR,
	author = {Wu, Haiyan and Qu, Yanyun and Lin, Shaohui and Zhou, Jian and Qiao, Ruizhi and Zhang, Zhizhong and Xie, Yuan and Ma, Lizhuang},
	year = {2021},
}

@inproceedings{Zheng23dehazecontrastive,
  author       = {Yu Zheng and
                  Jiahui Zhan and
                  Shengfeng He and
                  Junyu Dong and
                  Yong Du},
  title        = {Curricular Contrastive Regularization for Physics-Aware Single Image
                  Dehazing},
  booktitle =CVPR,
  pages        = {5785--5794},
  year         = {2023}
}

@article{PCL,
	title = {A practical contrastive learning framework for single-image super-resolution},

	journal = TNNLS,
	author = {Wu, Gang and Jiang, Junjun and Liu, Xianming},
  year={2024},
  volume={35},
  number={11},
  pages={15834-15845},
}

@inproceedings{Ran2024ilcr_contrastive_derain,
  author       = {Wu Ran and
                  Peirong Ma and
                  Zhiquan He and
                  Hao Ren and
                  Hong Lu},
  title        = {Harnessing Joint Rain-/Detail-aware Representations to Eliminate Intricate
                  Rains},
  booktitle       = ICLR,
  year         = {2024}
}

@inproceedings{Gao2024fadformer,
  author       = {Ning Gao and
                  Xingyu Jiang and
                  Xiuhui Zhang and
                  Yue Deng},
  title        = {Efficient Frequency-Domain Image Deraining with Contrastive Regularization},
booktitle = ECCV,
  pages        = {240--257},
  year         = {2024}
}

@inproceedings{wu2024learning,
	title = {Learning from history: Task-agnostic model contrastive learning for image restoration},
	booktitle = AAAI,
	author = {Wu, Gang and Jiang, Junjun and Jiang, Kui and Liu, Xianming},
	year = {2024},
pages={5976-5984}
}

@inproceedings{martin2001database1,
	title = {A database of human segmented natural images and its application to evaluating segmentation algorithms and measuring ecological statistics},
	volume = {2},
	pages = {416--423},
	booktitle = ICCV,
	author = {Martin, David R. and Fowlkes, Charless C. and Tal, Doron and Malik, Jitendra},
	year = {2001},
}

@inproceedings{yang2017deep,
	title = {Deep joint rain detection and removal from a single image},
	pages = {1685--1694},
	booktitle = CVPR,
	author = {Yang, Wenhan and Tan, Robby T. and Feng, Jiashi and Liu, Jiaying and Guo, Zongming and Yan, Shuicheng},
	year = {2017},
}

@inproceedings{Zamir_2021_CVPR_mprnet,
  author       = {Syed Waqas Zamir and
                  Aditya Arora and
                  Salman H. Khan and
                  Munawar Hayat and
                  Fahad Shahbaz Khan and
                  Ming{-}Hsuan Yang and
                  Ling Shao},
  title        = {Multi-Stage Progressive Image Restoration},
  booktitle    = CVPR,
  pages        = {14821--14831},
  year         = {2021}
}

@inproceedings{chen2023drsformer,
  author       = {Xiang Chen and
                  Hao Li and
                  Mingqiang Li and
                  Jinshan Pan},
  title        = {Learning {A} Sparse Transformer Network for Effective Image Deraining},
  booktitle    = CVPR,
  pages        = {5896--5905},
  year         = {2023}
}

@article{dl,
  author={Fan, Qingnan and Chen, Dongdong and Yuan, Lu and Hua, Gang and Yu, Nenghai and Chen, Baoquan},
  journal=TPAMI, 
  title={A General Decoupled Learning Framework for Parameterized Image Operators}, 
  year={2019}
}

@article{yao2024ndr,
  author       = {Mingde Yao and
                  Ruikang Xu and
                  Yuanshen Guan and
                  Jie Huang and
                  Zhiwei Xiong},
  title        = {Neural Degradation Representation Learning for All-in-One Image Restoration},
  journal      = TIP,
  volume       = {33},
  pages        = {5408--5423},
  year         = {2024}
}

@InProceedings{Conde2024InstructIR,
  author    = {Conde, Marcos V and Geigle, Gregor and Timofte, Radu},
  booktitle = ECCV,
  title     = {High-quality image restoration following human instructions},
  year      = {2024},
pages={1-12}
}

@inproceedings{nah2017deep,
	title = {Deep multi-scale convolutional neural network for dynamic scene deblurring},
	booktitle =CVPR,
	pages = {3883--3891},
	author = {Nah, Seungjun and Kim, Tae Hyun and Lee, Kyoung Mu},
	year = {2017-07},
}

@inproceedings{wei2018deep,
	title = {Deep retinex decomposition for low-light enhancement},
	pages = {155},
	booktitle =BMVC,
	author = {Wei, Chen and Wang, Wenjing and Yang, Wenhan and Liu, Jiaying},
	year = {2018},
}

@inproceedings{SwinIR,
	title = {{SwinIR}: Image restoration using swin transformer},
	pages = {1833--1844},
	booktitle = ICCV # { Workshops},
	author = {Liang, Jingyun and Cao, Jiezhang and Sun, Guolei and Zhang, Kai and Gool, Luc Van and Timofte, Radu},
	year = {2021},
}

@article{cui2024FSNet,
  author       = {Yuning Cui and
                  Wenqi Ren and
                  Xiaochun Cao and
                  Alois Knoll},
  title        = {Image Restoration via Frequency Selection},
  journal      = TPAMI,
  volume       = {46},
  number       = {2},
  pages        = {1093--1108},
  year         = {2024}
}

@inproceedings{TAPE,
  author       = {Lin Liu and
                  Lingxi Xie and
                  Xiaopeng Zhang and
                  Shanxin Yuan and
                  Xiangyu Chen and
                  Wengang Zhou and
                  Houqiang Li and
                  Qi Tian},
  title        = {{TAPE:} Task-Agnostic Prior Embedding for Image Restoration},
  booktitle    = ECCV,
  volume       = {13678},
  pages        = {447--464},
  year         = {2022}
}

@inproceedings{Transweather,
	title = {{TransWeather}: Transformer-based restoration of images degraded by adverse weather conditions},
	pages = {2343--2353},
	booktitle = CVPR,
	author = {Valanarasu, Jeya Maria Jose and Yasarla, Rajeev and Patel, Vishal M.},
	year = {2022},
}

@article{RESIDE_dataset,

title={Benchmarking Single-Image Dehazing and Beyond},

author={Li, Boyi and Ren, Wenqi and Fu, Dengpan and Tao, Dacheng and Feng, Dan and Zeng, Wenjun and Wang, Zhangyang},

journal=TIP,

volume={28},

number={1},

pages={492--505},

year={2019},
 

}

@article{Snow100K_dataset,
  author       = {Yun{-}Fu Liu and
                  Da{-}Wei Jaw and
                  Shih{-}Chia Huang and
                  Jenq{-}Neng Hwang},
  title        = {{DesnowNet}: Context-Aware Deep Network for Snow Removal},
  journal      = TIP,
  volume       = {27},
  number       = {6},
  pages        = {3064--3073},
  year         = {2018}
}

@inproceedings{weather_data,
  author       = {Yurui Zhu and
                  Tianyu Wang and
                  Xueyang Fu and
                  Xuanyu Yang and
                  Xin Guo and
                  Jifeng Dai and
                  Yu Qiao and
                  Xiaowei Hu},
  title        = {Learning Weather-General and Weather-Specific Features for Image Restoration
                  Under Multiple Adverse Weather Conditions},
  booktitle    = CVPR,
  pages        = {21747--21758},
  year         = {2023}
}

@inproceedings{raindrop,
  author       = {Rui Qian and
                  Robby T. Tan and
                  Wenhan Yang and
                  Jiajun Su and
                  Jiaying Liu},
  title        = {Attentive Generative Adversarial Network for Raindrop Removal From
                  a Single Image},
  booktitle    = CVPR,
  pages        = {2482--2491},
  year         = {2018}
}

@inproceedings{guo2024onerestore,
  title={{OneRestore}: A Universal Restoration Framework for Composite Degradation},
  author={Guo, Yu and Gao, Yuan and Lu, Yuxu and Liu, Ryan Wen and He, Shengfeng},
   booktitle    = ECCV,
  year={2024},
pages={255--272}
}

@inproceedings{SPANet_WangY0C0L19,
  author       = {Tianyu Wang and
                  Xin Yang and
                  Ke Xu and
                  Shaozhe Chen and
                  Qiang Zhang and
                  Rynson W. H. Lau},
  title        = {Spatial Attentive Single-Image Deraining With a High Quality Real
                  Rain Dataset},
  booktitle    = CVPR,
  pages        = {12270--12279},
  year         = {2019}
}

@inproceedings{outdoor,
  title={Heavy rain image restoration: Integrating physics model and conditional adversarial learning},
  author={Li, Ruoteng and Cheong, Loong-Fah and Tan, Robby T},
  booktitle=CVPR,
  pages={1633--1642},
  year={2019}
}

@article{xiao2022image,
	title = {Image de-raining transformer},
	journaltitle = {{IEEE} Transactions on Pattern Analysis and Machine Intelligence},
	journal = TPAMI,
	author = {Xiao, Jie and Fu, Xueyang and Liu, Aiping and Wu, Feng and Zha, Zheng-Jun},
	  year={2023},
  volume={45},
  number={11},
  pages={12978-12995},
}

@inproceedings{as2020,
  title={All in one bad weather removal using architectural search},
  author={Li, Ruoteng and Tan, Robby T and Cheong, Loong-Fah},
  booktitle=CVPR,
  pages={3175--3185},
  year={2020}
}

@Article{ozan2023weatherdiff,
  author  = {Ozan {\"{O}}zdenizci and Robert Legenstein},
  title   = {Restoring Vision in Adverse Weather Conditions With Patch-Based Denoising Diffusion Models},
  number  = {8},
  pages   = {10346--10357},
  volume  = {45},
  journal = TPAMI,
  year    = {2023},
}

@inproceedings{AWRCP_YeCBSXJYCL23,
  author       = {Tian Ye and
                  Sixiang Chen and
                  Jinbin Bai and
                  Jun Shi and
                  Chenghao Xue and
                  Jingxia Jiang and
                  Junjie Yin and
                  Erkang Chen and
                  Yun Liu},
  title        = {Adverse Weather Removal with Codebook Priors},
  booktitle    = ICCV,
  pages        = {12619--12630},
  year         = {2023}
}

@article{DehazeFormer,
  title={Vision Transformers for Single Image Dehazing},
  author={Song, Yuda and He, Zhuqing and Qian, Hui and Du, Xin},
  journal=TIP,
  year={2023},
  volume={32},
  pages={1927-1941}
}

@inproceedings{AiHZW024,
  author       = {Yuang Ai and
                  Huaibo Huang and
                  Xiaoqiang Zhou and
                  Jiexiang Wang and
                  Ran He},
  title        = {Multimodal Prompt Perceiver: Empower Adaptiveness, Generalizability
                  and Fidelity for All-in-One Image Restoration},
  booktitle = CVPR,
  pages        = {25432--25444},
  year         = {2024}
}

@article{Zhu2024MWFormer,
  author       = {Ruoxi Zhu and
                  Zhengzhong Tu and
                  Jiaming Liu and
                  Alan C. Bovik and
                  Yibo Fan},
  title        = {MWFormer: Multi-Weather Image Restoration Using Degradation-Aware
                  Transformers},
  journal      =TIP,
  volume       = {33},
  pages        = {6790--6805},
  year         = {2024}
}

@inproceedings{Zhang2024DCMPNet,
  author       = {Yafei Zhang and
                  Shen Zhou and
                  Huafeng Li},
  title        = {Depth Information Assisted Collaborative Mutual Promotion Network
                  for Single Image Dehazing},
  booktitle    = CVPR,
  pages        = {2846--2855},
  year         = {2024}
}

@inproceedings{Chen2024nerd-rain,
  author       = {Xiang Chen and
                  Jinshan Pan and
                  Jiangxin Dong},
  title        = {Bidirectional Multi-Scale Implicit Neural Representations for Image
                  Deraining},
  booktitle    = CVPR,
  pages        = {25627--25636},
  year         = {2024}
}

@article{Chen2022Snowformer,
  author       = {Sixiang Chen and
                  Tian Ye and
                  Yun Liu and
                  Erkang Chen and
                  Jun Shi and
                  Jingchun Zhou},
  title        = {SnowFormer: Scale-aware Transformer via Context Interaction for Single
                  Image Desnowing},
  journal      = {CoRR},
  volume       = {abs/2208.09703},
  year         = {2022}
}

@inproceedings{Zheng2024DiffUIR,
  author       = {Dian Zheng and
                  Xiao{-}Ming Wu and
                  Shuzhou Yang and
                  Jian Zhang and
                  Jian{-}Fang Hu and
                  Wei{-}Shi Zheng},
  title        = {Selective Hourglass Mapping for Universal Image Restoration Based
                  on Diffusion Model},
  booktitle    = CVPR,
  pages        = {25445--25455},
  year         = {2024}
}

@inproceedings{DTPM_0001CCXQL024,
  author       = {Tian Ye and
                  Sixiang Chen and
                  Wenhao Chai and
                  Zhaohu Xing and
                  Jing Qin and
                  Ge Lin and
                  Lei Zhu},
  title        = {Learning Diffusion Texture Priors for Image Restoration},
  booktitle    = CVPR,
  pages        = {2524--2534},
  year         = {2024}
}

@inproceedings{Zamfir2025MoCEIR,
  author       = {Eduard Zamfir and
                  Zongwei Wu and
                  Nancy Mehta and
                  Yuedong Tan and
                  Danda Pani Paudel and
                  Yulun Zhang and
                  Radu Timofte},
  title        = {Complexity Experts are Task-Discriminative Learners for Any Image
                  Restoration},
booktitle=CVPR, 
  pages        = {12753--12763},
  year         = {2025}
}

@inproceedings{Histoformer_SunRGWC24,
  author       = {Shangquan Sun and
                  Wenqi Ren and
                  Xinwei Gao and
                  Rui Wang and
                  Xiaochun Cao},
  title        = {Restoring Images in Adverse Weather Conditions via Histogram Transformer},
  booktitle    = ECCV,
  pages        = {111--129},
  year         = {2024}
}

@article{li2024latent,
  title={Latent diffusion enhanced rectangle transformer for hyperspectral image restoration},
  author={Li, Miaoyu and Fu, Ying and Zhang, Tao and Liu, Ji and Dou, Dejing and Yan, Chenggang and Zhang, Yulun},
  journal=TPAMI,
  year={2025},
  volume={47},
  number={1},
  pages={549--564},
  publisher={IEEE}
}

@article{wan2022old,
  title={Old photo restoration via deep latent space translation},
  author={Wan, Ziyu and Zhang, Bo and Chen, Dong and Zhang, Pan and Wen, Fang and Liao, Jing},
  journal=TPAMI,
  volume={45},
  number={2},
  pages={2071--2087},
  year={2022},
  publisher={IEEE}
}

@article{cui2024revitalizing,
  author={Cui, Yuning and Ren, Wenqi and Cao, Xiaochun and Knoll, Alois},
  journal={IEEE Transactions on Pattern Analysis and Machine Intelligence}, 
  title={Revitalizing Convolutional Network for Image Restoration}, 
  year={2024},
  volume={46},
  number={12},
  pages={9423-9438},
}

@ARTICLE{yue2025ResShift,
  author={Yue, Zongsheng and Wang, Jianyi and Loy, Chen Change},
  journal=TPAMI, 
  title={Efficient Diffusion Model for Image Restoration by Residual Shifting}, 
  year={2025},
  volume={47},
  number={1},
  pages={116-130},
}

@ARTICLE{Belhasin2024PUIR,
  author={Belhasin, Omer and Romano, Yaniv and Freedman, Daniel and Rivlin, Ehud and Elad, Michael},
  journal=TPAMI, 
  title={Principal Uncertainty Quantification With Spatial Correlation for Image Restoration Problems}, 
  year={2024},
  volume={46},
  number={5},
  pages={3321-3333},
}

@ARTICLE{Yue2024Difface,
  author={Yue, Zongsheng and Loy, Chen Change},
  journal=TPAMI, 
  title={DifFace: Blind Face Restoration With Diffused Error Contraction}, 
  year={2024},
  volume={46},
  number={12},
  pages={9991-10004},}

@ARTICLE{Zhang2025selfregression,
  author={Zhang, Zhao and Zhao, Suiyi and Jin, Xiaojie and Xu, Mingliang and Yang, Yi and Yan, Shuicheng and Wang, Meng},
  journal=TPAMI, 
  title={Noise Self-Regression: A New Learning Paradigm to Enhance Low-Light Images Without Task-Related Data}, 
  year={2025},
  volume={47},
  number={2},
  pages={1073-1088},
}

@ARTICLE{dai20247k++,
  author={Dai, Yuekun and Li, Chongyi and Zhou, Shangchen and Feng, Ruicheng and Luo, Yihang and Loy, Chen Change},
  journal=TPAMI, 
  title={Flare7K++: Mixing Synthetic and Real Datasets for Nighttime Flare Removal and Beyond}, 
  year={2024},
  volume={46},
  number={11},
  pages={7041-7055},
}

@ARTICLE{Chang2024derainingcontrastive,
  author={Chang, Yi and Guo, Yun and Ye, Yuntong and Yu, Changfeng and Zhu, Lin and Zhao, Xile and Yan, Luxin and Tian, Yonghong},
  journal=TPAMI, 
  title={Unsupervised Deraining: Where Asymmetric Contrastive Learning Meets Self-Similarity}, 
  year={2024},
  volume={46},
  number={5},
  pages={2638-2657},
}

@ARTICLE{Hsu2023waveletderain,
  author={Hsu, Wei-Yen and Chang, Wei-Chi},
  journal=TPAMI, 
  title={Wavelet Approximation-Aware Residual Network for Single Image Deraining}, 
  year={2023},
  volume={45},
  number={12},
  pages={15979-15995},
}

@ARTICLE{Feng2024realDehazing,
  author={Feng, Yuxin and Ma, Long and Meng, Xiaozhe and Zhou, Fan and Liu, Risheng and Su, Zhuo},
  journal=TPAMI, 
  title={Advancing Real-World Image Dehazing: Perspective, Modules, and Training}, 
  year={2024},
  volume={46},
  number={12},
  pages={9303-9320},
}

@ARTICLE{Quan2024GKMLdeblur,
  author={Quan, Yuhui and Wu, Zicong and Xu, Ruotao and Ji, Hui},
  journal=TPAMI, 
  title={Deep Single Image Defocus Deblurring via Gaussian Kernel Mixture Learning}, 
  year={2024},
  volume={46},
  number={12},
  pages={11361-11377},
}

@ARTICLE{Kim2022dehazeandlowlight,
  author={Kim, Guisik and Kwon, Junseok},
  journal=TITS, 
  title={Deep Illumination-Aware Dehazing With Low-Light and Detail Enhancement}, 
  year={2022},
  volume={23},
  number={3},
  pages={2494-2508},
  }

@INPROCEEDINGS{zheng2019derainanddenoise,
  author={Zheng, Qian and Shi, Boxin and Jiang, Xudong and Duan, Ling-Yu and Kot, Alex C.},
  booktitle=ICIP, 
  title={Denoising Adversarial Networks for Rain Removal and Reflection Removal}, 
  year={2019},
  volume={},
  number={},
  pages={2766-2770},
}

@inproceedings{cui2024hybridfrequency,
  author       = {Yuning Cui and
                  Mingyu Liu and
                  Wenqi Ren and
                  Alois Knoll},
  title        = {Hybrid Frequency Modulation Network for Image Restoration},
  booktitle    = IJCAI,
  pages        = {722--730},
  year         = {2024}
}

@inproceedings{Cui2023DualAttention,
  author       = {Yuning Cui and
                  Yi Tao and
                  Wenqi Ren and
                  Alois Knoll},
  title        = {Dual-Domain Attention for Image Deblurring},
  booktitle    = AAAI,
  pages        = {479--487},
  year         = {2023}
}

@inproceedings{Wang2024RFFNet,
  author       = {Qiang Wang and
                  Yuning Cui and
                  Yawen Li and
                  Yaping Ruan and
                  Ben Zhu and
                  Wenqi Ren},
  title        = {{RFFNet}: Towards Robust and Flexible Fusion for Low-Light Image Denoising},
  booktitle    = ACMMM,
  pages        = {836--845},
  year         = {2024}
}

@article{Cui2025EENet,
  author       = {Yuning Cui and
                  Qiang Wang and
                  Chaopeng Li and
                  Wenqi Ren and
                  Alois Knoll},
  title        = {{EENet}: An effective and efficient network for single image dehazing},
  journal      = {Pattern Recognit.},
  volume       = {158},
  pages        = {111074},
  year         = {2025}
}

@ARTICLE{Jiang2025DAWN,
  author={Jiang, Kui and Jiang, Junjun and Wang, Zheng and Geng, Zihan and Liu, Xianming},
  journal=TNNLS, 
  title={{DAWN+}: Wavelet-Based Image Deraining Meets Direction-Aware Attention and Mutual Representation}, 
  year={2025},
  volume={36},
  number={10},
  pages={18244-18258},
}

@ARTICLE{Jiang2021PCNet,
  author={Jiang, Kui and Wang, Zhongyuan and Yi, Peng and Chen, Chen and Wang, Zheng and Wang, Xiao and Jiang, Junjun and Lin, Chia-Wen},
  journal=TIP, 
  title={Rain-Free and Residue Hand-in-Hand: A Progressive Coupled Network for Real-Time Image Deraining}, 
  year={2021},
  volume={30},
  number={},
  pages={7404-7418},

}

\begin{IEEEbiography}[{\includegraphics[width=1.0in,height=1.25in,clip,keepaspectratio]{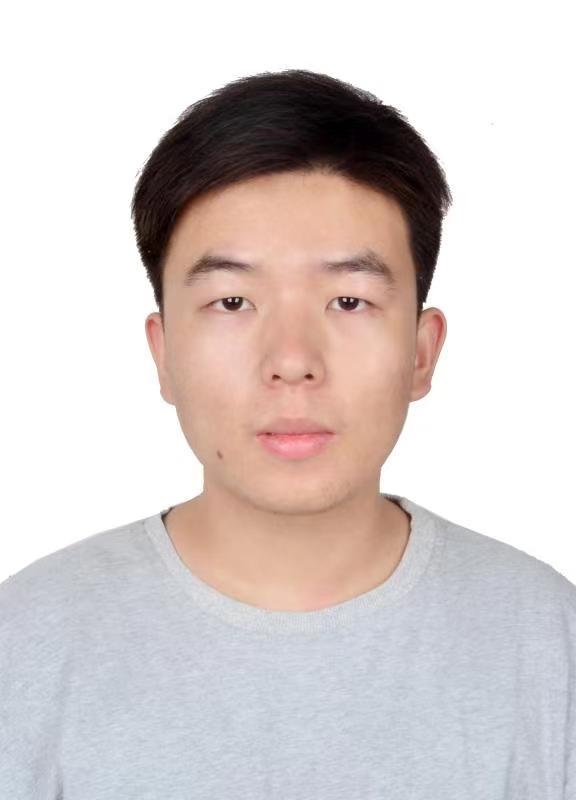}}]{Gang Wu} received the B.E. degree in the School of Computer Science and Technology from Soochow University, Jiangsu, China, in 2020. He is currently pursuing the Ph.D. degree in Faculty of Computing at Harbin Institute of Technology. His research interests include image restoration, representation learning, and self-supervised learning.
\vspace{-5mm}
\end{IEEEbiography}

\begin{IEEEbiography}[{\includegraphics[width=1.0in,height=1.25in,clip,keepaspectratio]{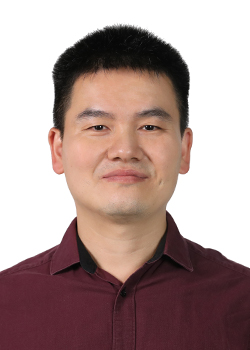}}]{Junjun Jiang}
received the B.S. degree in Mathematics from the Huaqiao University, Quanzhou, China, in 2009, and the Ph.D. degree in Computer Science from the Wuhan University, Wuhan, China, in 2014.

From 2015 to 2018, he was an Associate Professor with the School of Computer Science, China University of Geosciences, Wuhan. From 2016 to 2018, he was a Project Researcher with the National Institute of Informatics (NII), Tokyo, Japan. 
He is currently a Professor with the School of Computer Science and Technology, Harbin Institute of Technology, Harbin, China. His research interests include image processing and computer vision. 
\vspace{-5mm}

\end{IEEEbiography}

\begin{IEEEbiography}[{\includegraphics[width=1.0in,height=1.25in,clip,keepaspectratio]{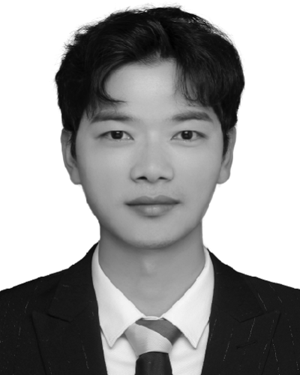}}]{Kui Jiang}
received the M.E. and Ph.D. degrees from the School of Computer Science, Wuhan University, Wuhan, China, in 2019 and 2022, respectively. Before July 2023, he was a Research Scientist with the Cloud BU, Huawei. He is currently an Associate Professor with the School of Computer Science and Technology, Harbin Institute of Technology. He received the 2022 ACM Wuhan Doctoral Dissertation Award. His research interests include image/video processing and computer vision.
\vspace{-5mm}

\end{IEEEbiography}

\begin{IEEEbiography}[{\includegraphics[width=1.0in,height=1.25in,clip,keepaspectratio]{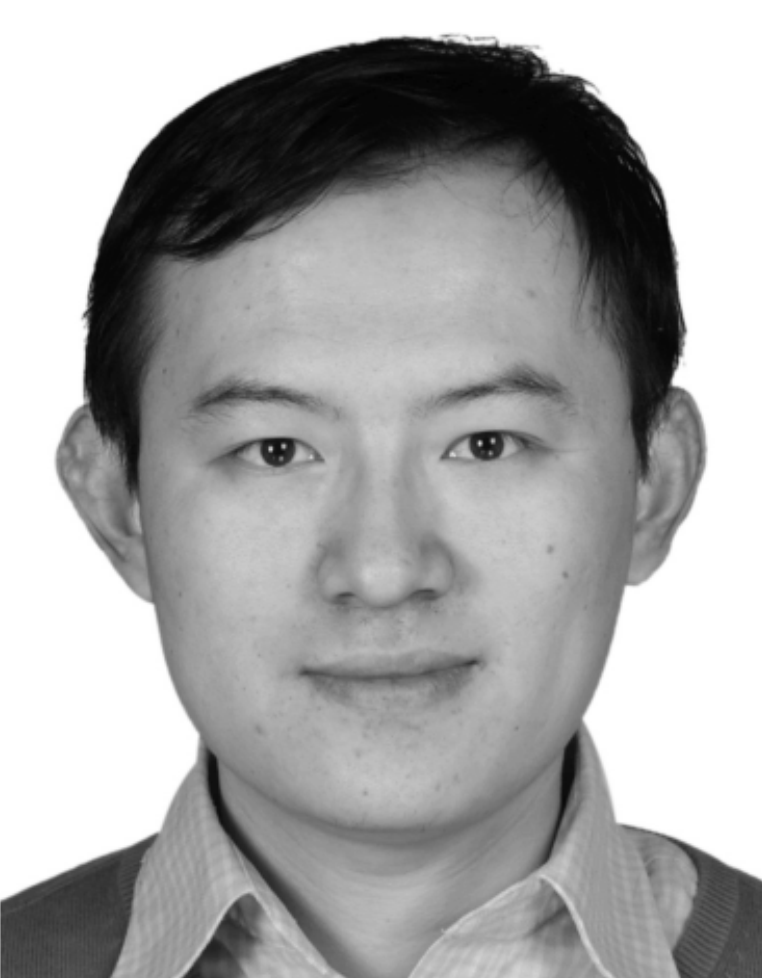}}]{Xianming Liu}
received the B.S., M.S., and Ph.D. degrees in computer science from the Harbin
Institute of Technology (HIT), Harbin, China,
in 2006, 2008, and 2012, respectively. In 2011, he spent half a year at the Department of Electrical and Computer Engineering, McMaster University, Canada, as a Visiting Student, where he was a Post-Doctoral Fellow from 2012 to 2013. He was a Project Researcher with the National Institute of Informatics (NII), Tokyo, Japan, from 2014 to 2017.
He is currently a Professor with the School of Computer Science and Technology, HIT. His research interests include trustworthy AI, 3D signal processing and biomedical signal processing.
\vspace{-5mm}

\end{IEEEbiography}

\begin{IEEEbiography}[{\includegraphics[width=1.0in,clip,keepaspectratio]{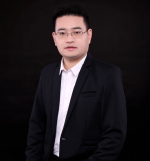}}]{Liqiang Nie}
received the B.Eng. and Ph.D. degree from Xi’an Jiaotong University and National University of Singapore (NUS), respectively. After Ph.D., he continued his research in NUS as a research fellow for three years. He is currently the dean of the Department of Computer Science and Technology, Harbin Institute of Technology (Shenzhen). His research interests lie primarily in multimedia computing and information retrieval. He is an AE of IEEE TIP, IEEE TKDE, IEEE TMM, IEEE TCSVT, ACM ToMM, and Information Science. Meanwhile, he serves as the chair of ICMR 2025, ICME 2025, and ACM MM 2027. He is a member of ICME steering committee. He has received many awards over the past four years, like SIGMM rising star in 2020, MIT TR35 China 2020, SIGIR best student paper in 2021, IEEE AI’s 10 to Watch in 2022, ACM MM Best paper award in 2022, and the National Youth Science and Technology Award in 2024.
\end{IEEEbiography}

\end{document}